\documentclass[journal]{IEEEtran}
\usepackage{cite}

\ifCLASSINFOpdf
\else
\fi
\usepackage{amsmath}
\usepackage{amssymb}
\ifCLASSOPTIONcompsoc
  \usepackage[caption=false,font=normalsize,labelfont=sf,textfont=sf]{subfig}
\else
  \usepackage[caption=false,font=footnotesize]{subfig}
\fi

\usepackage{stfloats}
\usepackage{glossaries}
\usepackage[]{algorithm2e}
\usepackage{tikz}
\usepackage{pgfplots}
\usepackage{siunitx}
\usepgfplotslibrary{groupplots}
\usetikzlibrary{arrows}
\usepackage{multirow}

\newacronym{QP}{QP}{Quadratic Problem}
\newacronym{QCQP}{QCQP}{Quadratically Constrained Quadratic Problem}
\newacronym{SQP}{SQP}{Sequential Quadratic Problem}
\newacronym{mpSQP}{mpSQP}{multi-parametric Sequential Quadratic Problem}
\newacronym{mpQP}{mpQP}{multi-parametric Quadratic Problems}
\newacronym{SOCP}{SOCP}{Second Order Conic Problem}
\newacronym{NLP}{NLP}{Nonlinear Problem}
\newacronym{OCP}{OCP}{Optimal Control Problem}
\newacronym{NOCP}{NOCP}{Nonlinear Optimal Control Problem}
\newacronym{MLTP}{MTLP}{Minimum Lap Time Problem}
\newacronym{NLMPP}{NLMPP}{Nonlinear Model Predictive Planner}
\newacronym{MPC}{MPC}{Model Predictive Control}
\newacronym{OSQP}{OSQP}{Operator Splitting Quadratic Problem}
\newacronym{IPOPT}{IPOPT}{Interior Point OPTimizer}
\newacronym{qpOASES}{qpOASES}{QP Online Active SEt Strategy}
\newacronym{ADMM}{ADMM}{Alternating Direction Method of Multipliers}
\newacronym{NDTM}{NDTM}{Nonlinear Double Track Model}
\newacronym{ECU}{ECU}{Electronic Control Unit}
\newacronym{RMSE}{RMSE}{Root Mean Square Error}
\newacronym{ES}{ES}{Energy Strategy}

\definecolor{TUMred}{RGB}{227,114,34}%
\definecolor{TUMgray}{rgb}{0.5977,0.5977,0.5977}
\definecolor{TUMgrayLight}{rgb}{0.8516,0.8398,0.7929}
\definecolor{TUMblue}{RGB}{0,101,189}%

\usetikzlibrary{external}
\usepackage{textcomp}
\usepackage{lipsum}
\newcommand\copyrighttext{%
	\footnotesize \textcopyright 2020 IEEE. Personal use of this material is permitted. Permission from IEEE must be obtained for all other uses, in any current or future media, including reprinting/republishing this material for advertising or promotional purposes, creating new collective works, for resale or redistribution to servers or lists, or reuse of any copyrighted component of this work in other works.
}
\newcommand\copyrightnotice{%
	\tikzset{external/export=false}
	\begin{tikzpicture}[remember picture,overlay]
		\node[anchor=south,yshift=6pt, xshift=0pt] at (current page.south) {\fbox{\parbox{\dimexpr\textwidth-\fboxsep-\fboxrule\relax}{\copyrighttext}}};
	\end{tikzpicture}%
	\tikzset{external/export=true}
}

\begin{document}
%
\title{Real-Time Adaptive Velocity Optimization for Autonomous Electric Cars at the Limits of Handling}
%
%
%

\author{Thomas~Herrmann,~%
		Alexander~Wischnewski,~%
		Leonhard~Hermansdorfer,~%
        Johannes~Betz,~%
        Markus~Lienkamp
\thanks{T. Herrmann (corresponding author), L. Hermansdorfer, J. Betz and M. Lienkamp are with the Institute of Automotive Technology, Department
of Mechanical Engineering, Technical University of Munich, Garching (Munich), 85748 Germany e-mail: thomas.herrmann@tum.de.}
\thanks{A. Wischnewski is with the Institute of Automatic Control, Department
	of Mechanical Engineering, Technical University of Munich, Garching (Munich), 85748 Germany}
\thanks{Manuscript received May 30, 2020; revised September 25, 2020; accepted December 24, 2020.}}

%
%

\markboth{IEEE Transactions on Intelligent Vehicles,~Vol.~XX, No.~X, Month~2021}%
{Shell \MakeLowercase{\textit{et al.}}: Bare Demo of IEEEtran.cls for IEEE Journals}
%



\maketitle

\copyrightnotice

\begin{abstract}
With the evolution of self-driving cars, autonomous racing series like Roborace and the Indy Autonomous Challenge are rapidly attracting growing attention. Researchers participating in these competitions hope to subsequently transfer their developed functionality to passenger vehicles, in order to improve self-driving technology for reasons of safety, and due to environmental and social benefits. The race track has the advantage of being a safe environment where challenging situations for the algorithms are permanently created. To achieve minimum lap times on the race track, it is important to gather and process information about external influences including, e.g., the
position of other cars and the friction potential between the road and the tires. Furthermore, the predicted behavior of the ego-car's propulsion system is crucial for leveraging the available energy as efficiently as possible. In this paper, we therefore present an optimization-based velocity planner, mathematically formulated as a multi-parametric Sequential Quadratic Problem (mpSQP). This planner can handle a spatially and temporally varying friction coefficient, and transfer a race Energy Strategy (ES) to the road. It further handles the velocity-profile-generation task for performance and emergency trajectories in real time on the vehicle's Electronic Control Unit (ECU).
\end{abstract}

\begin{IEEEkeywords}
Real-Time Numerical Optimization, Optimal Control, Velocity Planning, Trajectory Planning, Autonomous Electric Vehicles, Energy Strategy, Variable Friction Potential
\end{IEEEkeywords}

%
\IEEEpeerreviewmaketitle

\section{Introduction}
\label{sec:introduction}

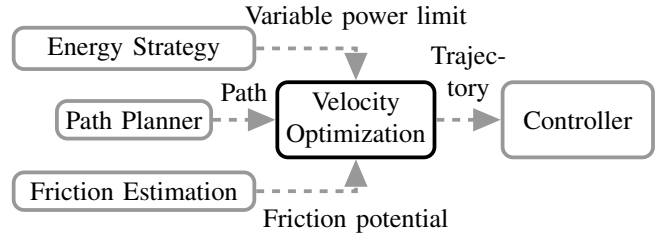
\begin{figure}[!tb]
	\centering
	\tikzstyle{EnvNode} = [rectangle, rounded corners, minimum width=3.2cm, minimum height=0.5cm,text centered, draw=TUMgray, line width=0.5mm]
\tikzstyle{SelfNode} = [rectangle, rounded corners, minimum width=2cm, minimum height=1.0cm, text centered, draw=black, line width=0.5mm]
\tikzstyle{arrow} = [->, -triangle 45, line width=0.5mm, black]
\tikzstyle{lineE} = [-, line width=0.5mm, TUMgray, dashed]
\tikzstyle{lineM} = [-, line width=1.5mm, black]
\begin{tikzpicture}[node distance=0.95cm]
\coordinate (Zero) at (0,0);
\node (ES) [EnvNode, align=center] at (Zero) {Energy Strategy};
\node (PP) [EnvNode, align=center, minimum width=2cm, below of=ES] {Path Planner};
\node (FE) [EnvNode, align=center, below of=PP] {Friction Estimation};
\node (VO) [SelfNode, align=center, right of=PP, xshift=2.0cm] {Velocity\\Optimization};
\node (CO) [EnvNode, align=center, right of=VO, minimum height=1.0cm, minimum width=2cm, xshift=2.0cm] {Controller};
%
%
\draw [arrow, TUMgray, dashed]  (ES.east) -| node [black,midway,above=0.1cm] {Variable power limit} (VO.north);
\draw [arrow, TUMgray, dashed]  (PP.east) -- node [black,midway,above=0.1cm] {Path} (VO.west);
\draw [arrow, TUMgray, dashed]  (FE.east) -| node [black,midway,below=0.1cm] {Friction potential} (VO.south);
\draw [arrow, TUMgray, dashed]  (VO.east) -- node [black,midway,above=0.1cm,align=center] {Trajec-\\tory} (CO.west);
\end{tikzpicture}
	\caption{Software environment of presented velocity optimization module.}
	\label{fig:architecture}
\end{figure}

\begin{figure}[!tb]
\centering
	\includegraphics[width=\columnwidth]{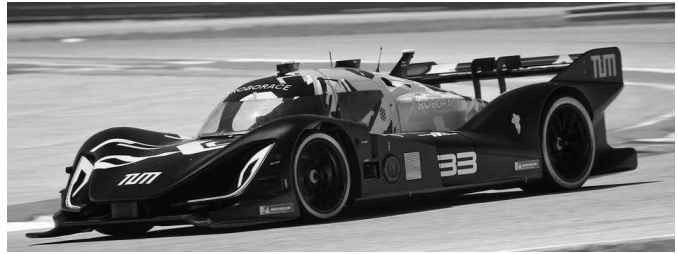}
\caption{TUM Roborace DevBot 2.0 on the race track.}
\label{fig:tum_devbot}
\end{figure}

%
%
%
%
\IEEEPARstart{T}{he} Technical University of Munich has been participating in the Roborace competition since 2018. Many parts of our software stack are already available on an open source basis \cite{ChairofAutomotiveTechnology2020} including the code of the algorithm in this work \cite{ChairofAutomotiveTechnology2020b}. This paper explains an optimization-based \gls{NLMPP}, mathematically formulated as a \gls{mpSQP} \cite{Morari2009}, to calculate the velocity profiles during a race. The velocity planner inputs are the offered race paths (``performance'' and ``emergency''), stemming from our graph-based path-planning framework \cite{Stahl2019}, see Fig. \ref{fig:architecture}. The presented velocity optimization in combination with the path framework span our local trajectory planner that will be used within the competition. The trajectory planner's output is forwarded to the underlying vehicle controller \cite{Heilmeier2019, Betz2019b}, transforming the target trajectory into actuator commands for the race car, called ``DevBot 2.0'', see Fig. \ref{fig:tum_devbot}. A huge motivation behind setting up an optimization-based velocity planner was to be able to handle information about the locally and temporally varying friction potential on the race track \cite{Hermansdorfer2019}, and utilize the information provided by the race \gls{ES} \cite{Herrmann2020} as a vehicle's velocity profile has a significant influence on it's energy consumption and power losses \cite{Ozatay2017}. The friction potential estimation and the calculation of the \gls{ES} are handled by separate modules in our software stack. Their outputs, the friction potential and variable power limits, are then considered in the presented velocity optimization algorithm.

To achieve real-time-capable calculation times, we build local approximations of the nonlinear velocity-planning problem, resulting in convex \gls{mpQP} \cite{Liniger2015} that can be solved iteratively using a \gls{SQP} method. We evaluated different open-source \gls{QP} solvers and compared their solution qualities and calculation times to a direct solution of the \gls{NLP}. We chose the \gls{OSQP} \cite{Stellato2020} solver as it outperformed its competitors on a standard x86-64 platform as well as on the DevBot's automotive-grade \gls{ECU}, the ARM-based NVIDIA Drive PX2.
\subsection{State of the art}
The field of trajectory planning of vehicles at the limits of handling is attracting growing attention in research. The scenarios where the car is required to be operated at the limits of its driving dynamics will become more and important as we see the spread of cars equipped with self-driving functionality, and even fully autonomous vehicles. Through this, complex scenarios with self-driving vehicles on the road will occur more frequently. Research is also being carried out on the race track where these challenging scenarios can deliberately be created in a safe environment \cite{Betz2019}.

In the field of global trajectory optimization for race tracks, different mathematical concepts are applied. In the work of Ebbesen et al. \cite{Ebbesen2018} a \gls{SOCP} formulation is used to calculate the optimal power distribution within the hybrid powertrain of a Formula One race car leading to globally time-optimal velocity profiles for a given path. For the same racing format, Limebeer and Perantoni \cite{Limebeer2015} took into account the 3D geometry of the race track within their formulation of an \gls{OCP} to solve a \gls{MLTP}. In a similar approach, Tremlett and Limebeer \cite{Tremlett2016} consider the thermodynamic effects of the tires. Christ et al. \cite{Christ2019} consider spatially variable but temporally fixed friction coefficients along the race track to calculate time-optimal global race trajectories for a sophisticated \gls{NDTM}. They show a significant influence of the variable friction coefficients on the achievable lap time when considered during the trajectory optimization process. A minimum-curvature \gls{QP} formulation, calculating the quasi-time-optimal trajectory for an autonomous race car on the basis of an occupancy grid map, is given by Heilmeier et al. \cite{Heilmeier2019}. Their advantage in comparison to \cite{Christ2019} is the computation time, but the resulting trajectories are suboptimal in terms of lap time. Also, Dal Bianco et al. \cite{DalBianco2018} formulate an \gls{OCP} to find the minimum lap time for a GP2 car and include a detailed multibody vehicle dynamics model with 14 degrees of freedom. However, none of these approaches are intended to work in real-time on a vehicle \gls{ECU}, but to deliver detailed and close-to-reality results for lap time or for the sensitivity analysis of vehicle setup parameters.

A further necessary and important part in a software stack for autonomous driving is the online re-planning of trajectories to avoid static and dynamic obstacles at high speeds. The literature can be structured into the three fields:
\begin{itemize}
	\item ``separated/two-step trajectory planning'', where the velocity calculation is a subsequent process of the path planner \cite{Huang2020, Meng2019, Zhang2018}.
	\item ``combined trajectory planning'', optimizing both path and velocity at the same time \cite{Subosits2019, Svensson2019, Mercy2018}.
	\item ``\gls{MPC} approaches'' taking into account the current vehicle state \cite{Liniger2015, Carvalho2013, Alrifaee2018, Williams2016, Alcala2020}.
\end{itemize}
In the following, we evaluate the literature according to the implemented features regarding
\begin{itemize}
	\item spatially and temporally varying friction coefficients,
	\item powertrain behavior,
	\item applicability at the limits of vehicle dynamics through fast computation times.
\end{itemize}
In the spline-based approach of Mercy et al. \cite{Mercy2018} trajectories for robots operating at low velocities are optimized. The calculation times of the general \gls{NLP}-solver \gls{IPOPT} \cite{Wachter2006} range up to several hundred milliseconds, which is too long for race car applications. Another general \gls{NLP}-formulation is done by Svensson et al. \cite{Svensson2018}. The latter describe a planning approach for safety trajectories of automated vehicles, which they validate experimentally in simulations for maximum velocities of \SI{30}{\kilo\meter\per\hour}, leveraging the general nonlinear optimal control toolkit ACADO \cite{Houska2011}.

Huang et al. \cite{Huang2020} describe a two-step approach: first determining the path across discretized available space and then calculating a sufficient velocity. Also, Meng et al. \cite{Meng2019} leverage a decoupled approach using a quadratic formulation for the speed profile optimization, reaching real-time capable calculation times below \SI{0.1}{\second} in this step. Nevertheless, both publications deal with low vehicle speeds in simulations of max. \SI{60}{\kilo\meter\per\hour}. Furthermore, Zhang et al. \cite{Zhang2018} implement a two-step algorithm where they use MTSOS \cite{Lipp2014} for the speed profile generation within several milliseconds for path lengths of up to \SI{100}{\meter}. The speed-profile optimization framework MTSOS developed by Lipp and Boyd \cite{Lipp2014} works for fixed paths leveraging a change of variables. As in the aforementioned publications, they consider a static friction coefficient and neglect the maximum available power of the car. The same is true for the \gls{MPC} algorithm by Carvalho et al. \cite{Carvalho2013}. They plan trajectories considering the driving dynamics of a bicycle model, neglecting physical constraints stemming from the powertrain, like maximum available torque or power. This is a major drawback for our application, as the DevBot 2.0 is often operating at the power limit of its electric machines.

Subosits and Gerdes \cite{Subosits2019} formulate a \gls{QCQP} replanning path and velocity of a race car at spatially fixed points on the track to avoid static obstacles. They consider a constant friction potential and the maximum available vehicle power. However, the obstacles need to be known in advance before the journey commences, and must be placed at a decent distance from the replanning points to allow the algorithm to find a feasible passing trajectory, given the physical constraints. In order to reach fast calculation times, Alrifaee et al. \cite{Alrifaee2018}  use a sequential linearization technique for real-time-capable trajectory optimization. They consider the friction maxima, with included velocity dependency that they determine beforehand. This dependency is assumed to be globally constant, thus neglecting the true track conditions during driving. Their experimental results stem from simulations with peak computation times of several hundred milliseconds on a desktop PC.

Considering variable friction on the road is attracting more attention, as it is an emergency-relevant feature for passenger cars and a performance-critical topic for race cars. Therefore, Svensson et al. \cite{Svensson2019} describe an adaptive trajectory-planning and optimization approach. They pre-sample trajectory primitives to avoid local optima in subsequent \gls{SQP}s stemming mainly from avoidance maneuvers to the suboptimal side of an obstacle. The vehicle adaptively reacts to a varying friction potential on the road at speeds of up to \SI{100}{\kilo\meter\per\hour}. The resulting problem is solved using simulations in MATLAB, so no information about the calculation speed on embedded hardware is given.

Stahl et al. \cite{Stahl2019} describe a two-step, multi-layered graph-based path planner. This approach allows for functionalities such as following other vehicles and overtaking maneuvers, also in non-convex scenarios at a high update rate. We use this path planner to generate the inputs for the velocity-optimization algorithm.

\subsection{Contributions}
In this paper we contribute to the state of the art in the field of real-time-capable trajectory planning with the following content.

(1) We formulate a tailored \gls{mpSQP} algorithm capable of adaptive velocity planning for race cars operating at the limits of handling, and at velocities above \SI{200}{\kilo\meter\per\hour}. The planner computes velocity profiles for various paths using the path planner \cite{Stahl2019} in real time on the target hardware, an NVIDIA Drive PX2 \cite{Denton2020} being an \gls{ECU} already proven for autonomous driving. The adaptivity refers to the multi-parametric input to the planner, depending on the vehicle's environment. The quadratic subproblems within the \gls{mpSQP} are handled using the \gls{OSQP} \cite{Stellato2020} solver. Its primal and dual infeasibility detection for convex problems \cite{Banjac2019} was integrated to flag up (as fast as possible) offered paths which could not feasibly be driven \cite{Stahl2019}.

(2) With the formulation of an \gls{mpSQP} optimization algorithm, it is possible to integrate our race \gls{ES}, described in our previous works \cite{Herrmann2020, Herrmann2019}. The necessary variable power parameters are forwarded to the velocity planner and considered as a hard constraint, see Fig. \ref{fig:architecture}. In the case of electric race cars, such an \gls{ES} is vital in order not to overstress the powertrain thermodynamically.

(3) We further allow the friction coefficient on the race track to vary spatially as well as temporally \cite{Hermansdorfer2019}. Therefore, global limits of the allowed longitudinal and lateral acceleration of the vehicle are omitted. This improves the achievable lap time significantly as the tires are locally exploited to their maximum. Via the temporal variation of the friction limits, we take into account varying grip due to, e.g., warming tires or changing weather conditions.

(4) To boost the solver selection for similar projects dealing with trajectory optimization within the community, we compare the efficiency of different solver types regarding calculation speed and solution quality. Therefore, we solve the quadratic subproblems in our \gls{mpSQP} using a first-order \gls{ADMM} implemented in the \gls{OSQP}-framework. Its results are compared to the active-set solver \gls{qpOASES} \cite{Ferreau2014}. We contrast both \gls{SQP}s with a direct solution of the nonlinear velocity optimization problem with the open-source, second-order interior point solver \gls{IPOPT} \cite{Wachter2006}, interfaced by the symbolic framework CasADi \cite{Andersson2019}.

Section \ref{sec:preliminaries} introduces the mathematical background of an \gls{SQP} method to solve an \gls{NLP}. In the following Section \ref{sec:optV}, the nonlinear equations of our velocity planner are introduced. We explain their efficient incorporation within an \gls{mpSQP} and explain details about the recursive feasibility of our optimization problem. The Results section shows the realization of the \gls{ES} and the handling of variable friction by our velocity-optimization algorithm. Furthermore, we contrast different solvers in terms of their runtime and solution quality.

\section{Preliminaries}
\label{sec:preliminaries}
In this section, the mathematical background to an \gls{SQP} optimization method to solve local approximations of an \gls{NLP} with objective function $J(\boldsymbol{o})$, $h_b(\boldsymbol{o})$ and $g_c(\boldsymbol{o})$, denoting equality and inequality constraints of scalar quantity $b$ and $c$, and optimization variables $\boldsymbol{o}$ is introduced.

The standard form of a \gls{NOCP} is given by \cite{Andersson2019}, \cite{Luenberger2008}:
\begin{align}	
\min~& J(\underbrace{x(s), u(s)}_{\boldsymbol{o}})\\
\mathrm{s.t.}~\frac{\mathrm{d}x(s)}{\mathrm{d}s} &= f(x(s), u(s))\\
h_b(\boldsymbol{o}) &= 0\\
g_c(\boldsymbol{o}) &\leq 0.
\label{eq:NLPP}
\end{align}
The independent space variable $s$ describes the distance along the vehicle's path in our problem. The function $f(x(s), u(s))$ specifies the derivatives of the state variable $x(s)$ as a function of the state $x(s)$ and the control input $u(s)$. 

The standard form of a \gls{QP} is expressed as \cite{Boyd2004}
\begin{align}
&\min{\quad\frac{1}{2} \boldsymbol{z}^T_\mathrm{qp} \boldsymbol{P} \boldsymbol{z}_\mathrm{qp} + \boldsymbol{q}^T \boldsymbol{z}_\mathrm{qp}} \nonumber \\
&\mathrm{s.t.} \quad \boldsymbol{l} \leq \boldsymbol{A}\boldsymbol{z}_\mathrm{qp} \leq \boldsymbol{u},
\label{eq:QP}
\end{align}
where $\boldsymbol{z}_\mathrm{qp}$ is the optimization vector, matrix $\boldsymbol{P}$ is the Hessian matrix of the discretized objective $J(\boldsymbol{o}^k)$ and the vector $\boldsymbol{q}^T$ equals the Jacobian of the discretized objective $\nabla J(\boldsymbol{o}^k)$ with iterate $k$. Matrix $\boldsymbol{A}$ contains the linearized versions of the constraints $h_b$ and $g_c$ in the optimization problem. Their upper and lower bounds are summarized in both vectors, $\boldsymbol{l}$ and $\boldsymbol{u}$.

In an \gls{SQP} method, the linearization point $\boldsymbol{o}^k$ is updated after every \gls{QP} iteration $k$ using \cite{Boggs1995}
\begin{align}
\boldsymbol{o}^{k+1} &= \boldsymbol{o}^k + \alpha \boldsymbol{z}_{\mathrm{qp}}%
\label{eq:zSQPUpdate} \\
\boldsymbol{\lambda}^{k+1} &= \boldsymbol{\lambda}^k_\mathrm{qp} \\
\boldsymbol{z}_{\mathrm{qp}} &= \boldsymbol{o} - \boldsymbol{o}^k.
\end{align}
In the quadratic subproblem, a solution for $\boldsymbol{z}_{\mathrm{qp}}$ is computed. We chose to initialize the Lagrange multiplier vector $\boldsymbol{\lambda}^{k+1}$ using the previous \gls{QP} solution as stated in the local SQP algorithm in \cite{Nocedal2006}.

On the one hand, the steplength parameter $\alpha$ must be calculated in order to perform a large step in the direction of the optimum $\boldsymbol{o}^*$ for fast convergence. On the other, $\alpha$ must be small enough to not skip or oscillate around $\boldsymbol{o}^*$. It is therefore necessary to define a suitable merit function, taking into account the minimization of the objective function as well as the adherence of the constraints \cite{Luenberger2008}, \cite{Boggs1995}. As it is hard to find such a merit function, we use the \gls{SQP} \gls{RMSE} $\bar{\varepsilon}_\mathrm{SQP}$ as well as the \gls{SQP} infinity norm error $\hat{\varepsilon}_\mathrm{SQP}$ to determine whether a stepsize $\alpha$ is suitable or not,
\begin{align}
\bar{\varepsilon}_\mathrm{SQP} &= \frac{1}{K}
\begin{Vmatrix}
\boldsymbol{o}^{k+1} - \boldsymbol{o}^k
\end{Vmatrix}_2 \leq \bar{\varepsilon}_\mathrm{SQP,tol} \label{eq:eps_sqp_bar}\\
\hat{\varepsilon}_\mathrm{SQP} &=
\begin{Vmatrix}
\boldsymbol{o}^{k+1} - \boldsymbol{o}^k
\end{Vmatrix}_\infty \leq \hat{\varepsilon}_\mathrm{SQP,tol},
\label{eq:eps_sqp_hat}
\end{align}
where $K$ denotes the number of elements in $\boldsymbol{o}^k$. In case one of the two errors $\varepsilon_\mathrm{SQP}$ increases when applying (\ref{eq:zSQPUpdate}), the counting variable $\gamma$ is increased and, therefore, $\alpha$ is reduced until the tolerance criteria values $\bar{\varepsilon}_\mathrm{SQP,tol}$ and $\hat{\varepsilon}_\mathrm{SQP,tol}$ are met:
\begin{align}
\alpha = \beta^{\gamma}.
\end{align}
The parameter $\beta \in \left]0, 1\right[$ is to be tuned problem-dependent as the Armijo rule states \cite{Luenberger2008} with $\gamma \in \left[0; 1; 2; ... \right]$.

To bring the objective $J(\boldsymbol{o})$ and the necessary nonlinear constraints $h_b(\boldsymbol{o})$ and $g_c(\boldsymbol{o})$ into the mathematical form of a \gls{QP} (\ref{eq:QP}), they are discretized and approximated quadratically or linearly, respectively, using Taylor series expansions in the form
\begin{align}
J(\boldsymbol{o}) \approx &\frac{1}{2} (\boldsymbol{o} - \boldsymbol{o}^k)^T \boldsymbol{P}(\boldsymbol{o}^k) (\boldsymbol{o} - \boldsymbol{o}^k) +\nonumber \\
&\nabla J(\boldsymbol{o}^k) (\boldsymbol{o} - \boldsymbol{o}^k) + J(\boldsymbol{o}^k)
\label{eq:J_approx}
\end{align}
and
\begin{align}
g_c(\boldsymbol{o}) \approx \nabla g(\boldsymbol{o}^k) (\boldsymbol{o} - \boldsymbol{o}^k) + g(\boldsymbol{o}^k).
\label{eq:linConstraints}
\end{align}

\section{Optimization-Based Velocity Planner}
\label{sec:optV}
This section describes the implemented point mass model, the used objective function, and the constraints necessary to optimize the velocity on the available paths. The point mass model was chosen, as it is commonly used to describe the driving dynamics in the automotive context. Due to its simplicity, it delivers a small number of optimization variables and constraints. Therefore, quick solver runtimes can be achieved. It still delivers quite accurate results for the task of pure velocity optimization \cite{Ebbesen2018}.

The concept of the optimization-based algorithm is to plan velocities with inputs from other software modules, cf. Section \ref{sec:introduction}. We do not deal with sensor noise in the planner but in the vehicle dynamics controller \cite{Heilmeier2019}, which receives the trajectory input. The trajectory planning module, consisting of a path planner \cite{Stahl2019} and the presented velocity optimization algorithm, always keeps the first discretization points of a new trajectory constant with the solution from a previous planning step. After matching the current vehicle position to the closest coordinate in the previously planned trajectory, a new plan starting from this position is made within the remaining part of a new trajectory. Through this, two control loops can be omitted, and prevent from unnecessary inferences in the planning and the control module.

\subsection{Nonlinear problem}
\label{subsec:NLP}
This subsection presents the nonlinear velocity optimization problem, structured into its system dynamics, equality and inequality constraints as well as its objective function. 
\subsubsection{System dynamics}
Let us first introduce the system dynamics of the point mass model for our physical vehicle state $v(s)$. Newton's second law for a point mass $m_\mathrm{v}$ states
\begin{equation}
m_\mathrm{v} v(s) \frac{\mathrm{d} v(s)}{\mathrm{d} s} = m_\mathrm{v} a_\mathrm{x}(s).
\end{equation}
With the derivative of the kinetic energy,
\begin{equation}
\frac{\mathrm{d}E_\mathrm{kin}(s)}{\mathrm{d} s} = \frac{1}{2} m_\mathrm{v} \frac{\mathrm{d}v^2(s)}{\mathrm{d}s} = m_\mathrm{v} v(s) \frac{\mathrm{d} v(s)}{\mathrm{d} s},
\end{equation}
the system dynamics are given by the longitudinal acceleration
\begin{equation}
a_\mathrm{x}(s) = \frac{1}{m_\mathrm{v}} \frac{\mathrm{d} E_\mathrm{kin}(s)}{\mathrm{d} s}.
\end{equation}
The force $F_\mathrm{x,p}(s)$ applied by the powertrain to move the point mass model can be calculated by
\begin{equation}
F_\mathrm{x,p}(s) = m_\mathrm{v} a_\mathrm{x}(s) + c_\mathrm{r} v^2(s)
\label{eq:FsAero}
\end{equation}
where $c_\mathrm{r}$ is the product of the air density $\rho_\mathrm{a}$, the air resistance coefficient $c_\mathrm{w}$ and the vehicle's frontal area $A_\mathrm{v}$,
\begin{equation}
c_\mathrm{r} = \frac{1}{2} \rho_\mathrm{a} c_\mathrm{w} A_\mathrm{v}.
\end{equation}

\subsubsection{Equality and inequality constraints}
To improve numerical stability and avoid backward movement, the velocity $v(s)$ is constrained,
\begin{equation}
0 \leq v(s) \leq v_\mathrm{max} (s).%
\label{eq:vmaxConstraint}
\end{equation}
To ensure that the optimization remains feasible in combination with a moving horizon, the terminal constraint
\begin{equation}
v(s_\mathrm{f}) \leq v_\mathrm{end}
\label{eq:vendConstraint}
\end{equation}
on the last coordinate point $s_\mathrm{f}$ within the optimization horizon is leveraged. Here, $v_\mathrm{end}$ denotes the minimal velocity the vehicle can take in the case of maximum specified track curvature $\kappa_\mathrm{max}$ at the vehicle's technically maximum possible lateral acceleration $a_\mathrm{y,max}$. Therefore,
\begin{equation}
v_\mathrm{end} = \sqrt{ \frac{a_\mathrm{y,max}}{\kappa_\mathrm{max}} }.
\label{eq:v_recursiveInfeas}
\end{equation}
At the beginning of the optimization horizon, the velocity and acceleration must equal the vehicle's target states of the currently executed plan $v_\mathrm{ini}$ and $a_\mathrm{x,ini}$,
\begin{align}
v(s_\mathrm{s}) &= v_\mathrm{ini},\nonumber\\
a_\mathrm{x,ini} - \delta_\mathrm{a} \leq a_\mathrm{x}(s_\mathrm{s}) &\leq a_\mathrm{x,ini} + \delta_\mathrm{a},
\end{align}
where $s_\mathrm{s}$ denotes the first coordinate within the moving optimization horizon and $\delta_\mathrm{a}$ a small tolerance to account for numerical imprecision.

As the vehicle's maximum braking as well as driving forces are technically limited, the resulting constraints are
\begin{equation}
F_\mathrm{min} \leq F_\mathrm{x,p}(s) \leq F_\mathrm{max}.
\end{equation}
The negative force constraint $F_\mathrm{min}$ does not affect the optimization-problem feasibility, as the DevBot's braking actuators can produce more negative force than the tires can transform.

The electric machine's output power $P(s)$ is computed using
\begin{equation}
P(s) = F_\mathrm{x,p}(s) v(s),
\end{equation}
limited by the available maximum
\begin{equation}
P(s) \leq P_\mathrm{max}(s).
\label{eq:EnergyConstraint}
\end{equation}
We highlight that $P_\mathrm{max}(s)$ is a space-dependent parameter in contrast to the constant maximum force $F_\mathrm{max}$. By this, the given race \gls{ES} based on our previous works \cite{Herrmann2020, Herrmann2019} is realized.

To further integrate the tire physics, we interpret the friction potential as a combined, diamond-shaped acceleration limit for the vehicle \cite{Hermansdorfer2019} given by the inequality
\begin{equation}
\begin{Vmatrix}
\left( \hat{a}_{\mathrm{x}}(s), \hat{a}_{\mathrm{y}}(s) \right)
\end{Vmatrix}_1 \leq 1 + \epsilon(s)
\label{eq:frictionConstraint}
\end{equation}
where $\left|\left| \cdot \right|\right|_1$ denotes the $l^1$-norm. Furthermore, the normalized longitudinal $\hat{a}_\mathrm{x}(s)$ as well as the lateral tire utilizations $\hat{a}_\mathrm{y}(s)$ are given by
\begin{equation}
\hat{a}_{\mathrm{x}}(s) = \frac{F_\mathrm{x,p}(s)}{m_\mathrm{v}}\frac{1}{\bar{a}_{\mathrm{x}}(s)}, \quad \hat{a}_{\mathrm{y}}(s) = \frac{a_{\mathrm{y}}(s)}{\bar{a}_{\mathrm{y}}(s)}.
\label{eq:var_accel_pars}
\end{equation}
Here, we use $\bar{a}_{\mathrm{x/y}}(s)$ to indicate a variable, space-dependent acceleration potential in both longitudinal and lateral direction, which is to be leveraged \cite{Hermansdorfer2019}. The lateral acceleration $a_\mathrm{y}(s)$ reads \cite{Braghin2008}
\begin{equation}
a_\mathrm{y}(s) = \kappa(s) v^2(s)
\label{eq:AccLat}
\end{equation}
accounting for the target path geometry by the variable road curvature parameter $\kappa(s)$.

In (\ref{eq:frictionConstraint}) the slack variable $\epsilon(s)$ ensures the recursive feasibility of the optimization problem: details are given in Subsection \ref{subsec:slackVariable}. We constrain the slack variable $\epsilon(s)$ by
\begin{equation}
0 \leq \epsilon(s) \leq \epsilon_\mathrm{max}
\label{eq:EpsilonConstraint}
\end{equation}
to prohibit negative values and additionally keep the physical tire exploitation within a specified maximum.

Similarly to (\ref{eq:v_recursiveInfeas}), the longitudinal and lateral acceleration limits at the end of the optimization horizon $\bar{a}_\mathrm{x}(s_\mathrm{f})$ and $\bar{a}_\mathrm{y}(s_\mathrm{f})$ must be set to the lowest physically possible acceleration limits $\bar{a}_\mathrm{x/y,min}$ for the current track conditions,
\begin{equation}
\bar{a}_\mathrm{x/y}(s_\mathrm{f}) \leq \bar{a}_\mathrm{x/y,min}.
\label{eq:acc_recursive}
\end{equation}

\subsubsection{Continuous objective function}
With the help of the introduced symbols and equations we can now formulate the objective function $J(x(s))$ to minimize the traveling time along the given path:
\begin{align}
	J(x(s)) =& \int_{0}^{s_\mathrm{f}}{\frac{1}{v(s)}\mathrm{d}s} +
	\frac{\rho_{\mathrm{j}}}{s_\mathrm{f}}\int_{0}^{s_\mathrm{f}}{\left( \frac{\mathrm{d}^2 v(s)}{\mathrm{d}^2 t}\right)^2} \mathrm{d}s +\nonumber\\ 
	& \frac{\rho_{\mathrm{\epsilon,l}}}{s_\mathrm{f}} \int_{0}^{s_\mathrm{f}}{\epsilon(s)}\mathrm{d}s  + \frac{\rho_{\mathrm{\epsilon,q}}}{s_\mathrm{f}} \int_{0}^{s_\mathrm{f}}{\epsilon^2(s)}\mathrm{d}s.
	\label{eq:nonlinearObjective}
\end{align}
We chose the optimization variables $\boldsymbol{o}$ to be the state velocity $v(s)$ as well as the slacks $\epsilon(s)$. The control input to the vehicle $u(s) = F_\mathrm{x,p}(s)$ doesn't occur explicitly in the objective function but can be recalculated from the state trajectory $v(s)$, cf. (\ref{eq:FsAero}).

Minimizing the term $\frac{1}{v(s)}$ is equivalent to the minimization of the lethargy $\frac{\mathrm{d}t}{\mathrm{d}s}$, which can be interpreted as the time necessary to drive a unit distance \cite{Ebbesen2018}. To weight the different terms, the penalty parameters $\rho$ are used. These include a jerk penalty $\rho_{\mathrm{j}}$, a slack weight $\rho_{\mathrm{\epsilon,l}}$ on their integral and a penalty $\rho_{\mathrm{\epsilon,q}}$ on the integral of the squared slack values. The linear penalty term on the slack variable $\epsilon(s)$ is necessary to achieve an exact penalty maintaining the original problem's optimum $\left[v^*(s)~\epsilon^*(s)\right]$ if feasible \cite{Kerrigan2000}. Similar to a regularization term, the integral of the squared slacks $\epsilon(s)$ is additionally added to improve numerical stability and the smoothness of the results.
\subsection{Multi-parametric Sequential Quadratic Problem}
\label{subsec:efficient_implementation}
This chapter gives details about the implementation of the \gls{NLP} given in Subsection \ref{subsec:NLP} as an \gls{mpSQP} in order to efficiently solve local approximations of the velocity planning problem. We describe how to approximate the nonlinear objective function $J(x(s))$ (\ref{eq:nonlinearObjective}) to achieve a constant and tuneable Hessian matrix within our tailored \gls{mpSQP} algorithm. Furthermore, we present a method to reduce the number of slack variables $\epsilon(s)$ and the slack constraints (\ref{eq:EpsilonConstraint}) therefore necessary.

Our optimization vector $\boldsymbol{z}=\boldsymbol{o}^k$ in a discrete formulation transforms into
\begin{align}
\boldsymbol{z} =
&\begin{bmatrix}
\underbrace{v_1(s_1) \ldots v_{M - 1}(s_{M-1})}_{\boldsymbol{v}} & \underbrace{\epsilon_0(s_0) \ldots \epsilon_{N - 1}(s_{N - 1})}_{\boldsymbol{\epsilon}}
\end{bmatrix}^T \\
&\in \mathbb{R}^{K \times 1} \nonumber
\end{align}
where $K = M - 1 + N$ where $M$ denotes the number of discrete velocity points $v_m$ and $N$ the number of discrete slack variables $\epsilon_n$ used in the tire inequality constraints within one optimization horizon. We drop the dependency of $\boldsymbol{z}$ on $s_m$ in the following for the sake of readability. The velocity variable $v_0$ is removed from the vector $\boldsymbol{z}$ as it is a fixed parameter equaling the velocity planned in a previous \gls{SQP} $l - 1$ for the current position.

To reduce the problem size, we apply one slack variable $\epsilon_n$ to multiple consecutive discrete velocity points $v_m$. This is done uniformly and leads to:  
\begin{align}
	\begin{bmatrix}
		\underbrace{
		v_1~\ldots~v_{\tilde{N}}}_{\epsilon_0}%
		& \underbrace{
		v_{\tilde{N} + 1}~\ldots~v_{2\tilde{N}}}_{\epsilon_1}%
		& \underbrace{v_{2\tilde{N} + 1}~\ldots~v_{M - 1}}_{\ldots}
		\label{eq:slackVisualization}
	\end{bmatrix}
\end{align}
Here, $\tilde{N}$ is a problem-specific parameter setting a trade off between the number of optimization variables and therefore the calculation speed and accuracy in the solution.

From domain knowledge we know that the objective function can be approximated in the form
\begin{align}
J(\boldsymbol{z}) \approx
&\underbrace{
\begin{Vmatrix}
\boldsymbol{v} - \boldsymbol{v}_\mathrm{max}
\end{Vmatrix}^2_2}_{J_\mathrm{v}} +%
\underbrace{\rho_{\mathrm{j}} \begin{Vmatrix} \Delta \boldsymbol{v} \end{Vmatrix}^2_2}_{J_\mathrm{j}} + \nonumber\\
&
\underbrace{
	\rho_{\mathrm{\epsilon,l}} \begin{Vmatrix} \zeta \boldsymbol{\epsilon} \end{Vmatrix}^2_1}_{J_\mathrm{\epsilon,l}} +
\underbrace{
\rho_{\mathrm{\epsilon,q}} \begin{Vmatrix} \zeta \boldsymbol{\epsilon} \end{Vmatrix}^2_2}_{J_\mathrm{\epsilon,q}}.
\label{eq:discreteObjective}
\end{align}
The slack variables are transformed via the constant factor $\zeta$; how this is selected is discussed at the end of this section. By using the $l^2$-norm of the vector difference of $\boldsymbol{v}$ and $\boldsymbol{v}_\mathrm{max}$, the solution tends to minimize the travel time along the path. Still, this formulation in combination  with (\ref{eq:vmaxConstraint}) makes the car keep a specified maximum velocity $v_\mathrm{max}(s)$ dependent on the current position $s$ to react, e.g., to other cars. To control the vehicle's jerk behavior, we add the Tikhonov regularization term $\rho_{\mathrm{j}} \begin{Vmatrix} \Delta \boldsymbol{v} \end{Vmatrix}^2_2$ \cite{Boyd2004} that approximates the second derivative of $\boldsymbol{v}$. The tridiagonal Toeplitz matrix $\Delta\in \mathbb{R}^{M - 3 \times M - 1}$ contains the diagonal elements $\left(\begin{smallmatrix}1 & -2 & 1\end{smallmatrix}\right)$ \cite{Boyd2004}. By the $l^1$-norm within $J_\mathrm{\epsilon,l}$, the summation of the absolute values of the slack variable vector entries in $\boldsymbol{\epsilon}$ is achieved. To improve the numerical conditioning of the problem, their $l^2$-norm is added additionally by $J_\mathrm{\epsilon,q}$.

For the specific choice of cost function in (\ref{eq:J_approx}), the Hessian matrix $\boldsymbol{P} \in \mathbb{R}^{K \times K}$ does not depend on $\boldsymbol{z}$. The condition number $\sigma_\mathrm{H}$ of the Hessian $\boldsymbol{P}$ is tuned to be as close to 1 as possible via the penalties $\rho_{\mathrm{j}}$ and $\rho_{\mathrm{\epsilon,l}}$ as well as $\zeta$ denoting the unit conversion factor of the tire slack variable values in $\boldsymbol{\epsilon}$ to SI units:
\begin{equation}
\renewcommand{\arraystretch}{1.1}
\boldsymbol{P} = 
\left[
\begin{array}{ccc|cc}
\ddots & \ddots & 0 & & \\
\ddots & c_j\rho_\mathrm{j} & \ddots & \multicolumn{2}{c}{0} \\
0 & \ddots & \ddots & & \\
\hline
\multicolumn{3}{c|}{\multirow{2}{*}{0}} & 2\rho_{\mathrm{\epsilon,q}} \zeta^2 & 0\\
& & & 0 & \ddots\\
\end{array}
\right]
\label{eq:Hess}
\end{equation}
The function $c_j \rho_\mathrm{j}$ represents different constant entries $j$ that are linearly dependent on $\rho_\mathrm{j}$. This upper left part of $\boldsymbol{P}$ is a bisymmetric matrix with constant entries on its main diagonal, as well as on its first and second ones.

By using the approach of multi-parametric programming, we can vary several problem parameters online in the \gls{SQP} (Section \ref{sec:preliminaries}) without changing the problem size. These parameters include the
\begin{itemize}
	\item spatial discretization length $\Delta s_m$.
	\item curvature of the local path $\kappa (s_m)$ \cite{Stahl2019}.
	\item maximum allowed velocity $v_\mathrm{max} (s_m)$.
	\item power limitations $P_\mathrm{max} (s_m)$ stemming from a global race strategy taking energy limitations into account \cite{Herrmann2020, Herrmann2019}.
	\item longitudinal and lateral acceleration limits $\bar{a}_\mathrm{x}(s_m)$, $\bar{a}_\mathrm{y}(s_m)$ \cite{Hermansdorfer2019}.
\end{itemize}

\subsection{Variable acceleration limits}
\label{sec:variable_friction}
To fully utilize the maximum possible tire forces, a time- and location-dependent map of the race track, containing the maximum possible accelerations, is generated. The acceleration limits can be interpreted as vehicle-related friction coefficients, cf. Subsection \ref{subsec:NLP}. The 1D map along the global coordinate $s_\mathrm{glo}$ with variable discretization step length stores the individual acceleration limitations $\Sigma_{\bar{a}}(s_\mathrm{glo})$ in longitudinal and lateral directions. The acceleration limits are used in the selected local path as the parameters $\bar{a}_\mathrm{x/y}(s)$ (\ref{eq:var_accel_pars}). It is important to know that while the vehicle proceeds, the target path is updated constantly, i.e., the global coordinates $s_\mathrm{glo}$ selected as the local path $s$ vary permanently in subsequent velocity optimizations. The path planning guarantees to re-use the first few global coordinates $s_\mathrm{glo}$ from a previous timestep $t^0$ as the starting coordinates $s_m$ of the subsequently chosen path at $t^1$. Still, the path coordinates $s_m$ at the end of the planning horizon are not guaranteed to precisely match all of the previously used global coordinates $s_\mathrm{glo}$ since the path might change. Therefore, the requested coordinates in the acceleration map do also vary slightly, but are matched to the same local coordinate indices in $s_m$ within the local path in subsequent timesteps. This leads to differences in the acceleration limits $\bar{a}_\mathrm{x/y}(s_m)$ in subsequent planning iterations for identical local path indices $m$ and therefore probably to infeasible problems in terms of optimization, see Section \ref{subsec:slackVariable}.

\begin{figure}[!tb]
	\centering
	\resizebox{1.025\columnwidth}{!}{
		\begin{tikzpicture}

\begin{groupplot}[group style={group size=1 by 2, vertical sep=15pt},
	width=8cm,
	height=2.5cm,
	scale only axis,
	xmajorgrids]
\nextgroupplot[
legend columns=3,
legend cell align={left},
legend style={draw=none, at={(0.5, 1.45)}, anchor=north, /tikz/every even column/.append style={column sep=0.15cm}},
legend entries={path coord. at $t^0$,
				path coord. at $t^1$,
				$\Sigma_{\bar{a}}(s_\mathrm{glo})$,
				$\bar{a}_\mathrm{x/y}(s_m)$ at $t^0$,
				$\bar{a}_\mathrm{x/y}(s_m)$ at $t^1$,
				$\tilde{\Sigma}_{\bar{a}}(s_\mathrm{glo})$},
tick align=outside,
xmin=0, xmax=70,
xtick={0,10,20,30,40,50,60,70},
xticklabels={,,},
ylabel style={align=center},
ylabel={Acc. limits\\$\bar{a}_\mathrm{x/y}$ in \SI{}{\meter\per\square\second}},
ymin=9.5, ymax=13.5,
ytick={10,11,12,13},
axis x line*=bottom,
axis y line*=left,
]
\addplot [only marks, mark=x, mark size=3pt, draw=TUMblue, fill=TUMblue]
table{%
0 11
5.5 11
11 12
16.5 12
22 12.5
27.5 12.5
33 13
38.5 13
44 12
49.5 12
55 11
60.5 10
66 10
71.5 12
77 12
};
\addplot [only marks, mark=*, draw=TUMred, fill=TUMred]
table{%
2 11
7.5 11
13 12
18.5 12
24 12.5
29.5 12.5
35 13
40.5 12
46 12
51.5 11
57 11
62.5 10
68 10
73.5 12
79 12
};
\path [fill=TUMgray, opacity=0.5]
(axis cs:11,11)
--(axis cs:11,12)
--(axis cs:13,12)
--(axis cs:13,11)
--cycle

(axis cs:22,12)
--(axis cs:22,12.5)
--(axis cs:24,12.5)
--(axis cs:24,12)
--cycle

(axis cs:33,12.5)
--(axis cs:33,13)
--(axis cs:35,13)
--(axis cs:35,12.5)
--cycle

(axis cs:40.5,13)
--(axis cs:44,13)
--(axis cs:44,12)
--(axis cs:40.5,12)
--cycle

(axis cs:51.5,12)
--(axis cs:55,12)
--(axis cs:55,11)
--(axis cs:51.5,11)
--cycle;

\addplot [thick, black, const plot mark left]
table {%
0 11
10 12
20 12.5
30 13
40 12
50 11
60 10
70 10
};
\addplot [semithick, TUMblue, const plot mark left]
table {%
0 11
5.5 11
11 12
16.5 12
22 12.5
27.5 12.5
33 13
38.5 13
44 12
49.5 12
55 11
60.5 10
66 10
71.5 12
77 12
};
\addplot [semithick, TUMred, const plot mark left]
table {%
2 11
7.5 11
13 12
18.5 12
24 12.5
29.5 12.5
35 13
40.5 12
46 12
51.5 11
57 11
62.5 10
68 10
73.5 12
79 12
};

\addlegendimage{black, thick, dashed}

\nextgroupplot[
tick align=outside,
xlabel={Global coordinate $s_\mathrm{glo}$ in \SI{}{\meter}},
xmin=0, xmax=70,
xtick={0,10,20,30,40,50,60,70},
xticklabels={200,210,220,230,240,250,260,270},
ylabel style={align=center},
ylabel={Acc. limits\\$\bar{a}_\mathrm{x/y}$ in \SI{}{\meter\per\square\second}},
ymin=9.5, ymax=13.5,
ytick={10,11,12,13},
axis x line*=bottom,
axis y line*=left,
]
\addplot [only marks, mark=x, mark size=3pt, draw=TUMblue, fill=TUMblue]
table{%
0 11
5.5 11
11 11.1
16.5 11.65
22 12.1
27.5 12.375
33 12.35
38.5 12.075
44 11.6
49.5 11.05
55 10.5
60.5 10
66 10
71.5 10
77 10
};
\addplot [only marks, mark=*, draw=TUMred, fill=TUMred]
table{%
2 11
7.5 11
13 11.3
18.5 11.85
24 12.2
29.5 12.475
35 12.25
40.5 11.95
46 11.4
51.5 10.85
57 10.3
62.5 10
68 10
73.5 10
79 10
};
\addplot [thick, black, const plot mark left]
table {%
0 11
10 12
20 12.5
30 13
40 12
50 11
60 10
70 10
};
\addplot [thick, black, dashed]
table {%
0 11
10 11
20 12
30 12.5
40 12
50 11
60 10
70 10
};
\addplot [semithick, TUMblue, const plot mark left]
table {%
0 11
5.5 11
11 11.1
16.5 11.65
22 12.1
27.5 12.375
33 12.35
38.5 12.075
44 11.6
49.5 11.05
55 10.5
60.5 10
66 10
71.5 10
77 10
};
\addplot [semithick, TUMred, const plot mark left]
table {%
2 11
7.5 11
13 11.3
18.5 11.85
24 12.2
29.5 12.475
35 12.25
40.5 11.95
46 11.4
51.5 10.85
57 10.3
62.5 10
68 10
73.5 10
79 10
};

\path [fill=TUMgray, opacity=0.5]
(axis cs:11,11)
--(axis cs:11,11.1)
--(axis cs:13,11.1)
--(axis cs:13,11)
--cycle

(axis cs:16.5,11.3)
--(axis cs:16.5,11.65)
--(axis cs:18.5,11.65)
--(axis cs:18.5,11.3)
--cycle

(axis cs:22,11.85)
--(axis cs:22,12.1)
--(axis cs:24,12.1)
--(axis cs:24,11.85)
--cycle

(axis cs:27.5,12.2)
--(axis cs:27.5,12.375)
--(axis cs:29.5,12.375)
--(axis cs:29.5,12.2)
--cycle

(axis cs:35,12.25)
--(axis cs:35,12.35)
--(axis cs:38.5,12.35)
--(axis cs:38.5,12.25)
--cycle

(axis cs:40.5,11.95)
--(axis cs:40.5,12.075)
--(axis cs:44,12.075)
--(axis cs:44,11.95)
--cycle

(axis cs:46,11.4)
--(axis cs:46,11.6)
--(axis cs:49.5,11.6)
--(axis cs:49.5,11.4)
--cycle

(axis cs:51.5,10.85)
--(axis cs:51.5,11.05)
--(axis cs:55,11.05)
--(axis cs:55,10.85)
--cycle

(axis cs:57,10.3)
--(axis cs:57,10.5)
--(axis cs:60.5,10.5)
--(axis cs:60.5,10.3)
--cycle;
\end{groupplot}

\end{tikzpicture}}
	\caption{Diagram of acceleration limits for two subsequently planned paths; with pure readout values $\Sigma_{\bar{a}}(s_\mathrm{glo})$ (top) and with the proposed interpolation scheme $\tilde{\Sigma}_{\bar{a}}(s_\mathrm{glo})$ (bottom). Gray areas show where the subsequently planned path receives decreased maximum acceleration limits $\bar{a}_\mathrm{x/y}(s_m)$ due to tolerances in the spatial discretization.}
	\label{fig:map_discretization}
\end{figure}
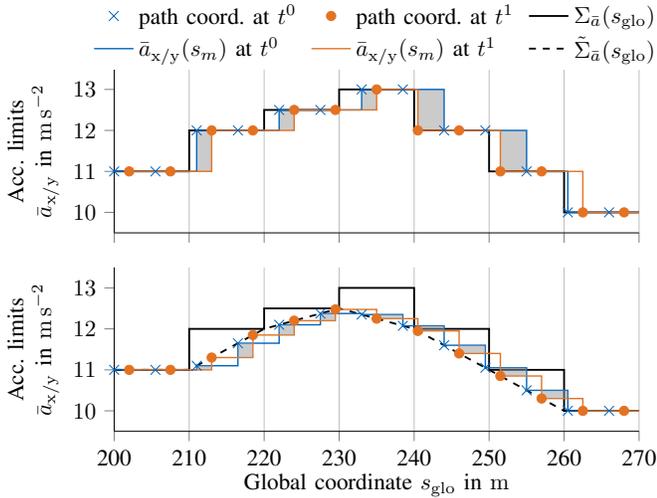

The nature of the discretization problem is illustrated in Fig.~\ref{fig:map_discretization}. The given example provides stored acceleration limits $\Sigma_{\bar{a}}(s_\mathrm{glo})$ in \SI{10}{\meter} steps. The planning horizon ranges from $s_\mathrm{glo}$ = \SIrange{0}{300}{\meter}. Therefore, we show a snippet of the end of the planning horizon ($s_\mathrm{glo}$ = \SIrange{200}{270}{\meter}) as the discretization issues are clearly visible here. At timestep $t^0$, the planning algorithm  requests the stored acceleration limits $\Sigma_{\bar{a}}(s_\mathrm{glo})$ every \SI{5.5}{\meter}, starting at $s_\mathrm{glo}=\SI{0}{\meter}$. Within the depicted path snippet, the subsequent iteration at $t^1$ starts at a shift of \SI{2.0}{\meter} with the same stepsize.

The simple approach of directly obtaining the local acceleration limits from the stored values $\Sigma_{\bar{a}}(s_\mathrm{glo})$ by applying zero-order hold comes with drawbacks. A slight shift in the global coordinate selection can lead to situations where the acceleration limits $\bar{a}_\mathrm{x/y}(s_m)$ differ between subsequent timesteps ($t^0$, $t^1$) in the local path. If the subsequent acceleration limits $\bar{a}_\mathrm{x/y}(s_m)$ at $t^1$ are smaller, they can lead to infeasibility. In Fig. \ref{fig:map_discretization}, the gray areas highlight the situations where the obtained acceleration limits $\bar{a}_\mathrm{x/y}(s_m)$ at $t^1$ are smaller compared to $t^0$ for identical local path indices $m$.

To mitigate the discretization effects, we propose an interpolation scheme leading to the values $\tilde{\Sigma}_{\bar{a}}(s_\mathrm{glo})$. It applies linear interpolation between the stored acceleration limits $\Sigma_{\bar{a}}(s_\mathrm{glo})$ but acts cautiously in the sense that it always underestimates the actual values $\Sigma_{\bar{a}}(s_\mathrm{glo})$. This can be seen from $s_\mathrm{glo}$ =  \SIrange{200}{210}{\meter}, where the value is kept constant instead of interpolating between \SI{11}{} and \SI{12}{\meter\per\square\second}, and from $s_\mathrm{glo}$ = \SIrange{230}{240}{\meter} where the algorithm adapts to the decreasing values although the stored value is \SI{13}{\meter\per\square\second}. Then, zero-order hold is applied to this conservatively interpolated line. The limits obtained at $t^1$ often lie above $\bar{a}_\mathrm{x/y}(s_m)$ at $t^0$, which allow higher accelerations $a_\mathrm{x/y}(s_m)$ than expected at $t^0$ and thus compensating for overestimated areas~(gray areas).

The acceleration limits $\Sigma_{\bar{a}}(s_\mathrm{glo})$ are constantly updated by an estimation algorithm \cite{Hermansdorfer2019} and are therefore also considered time-variant. During the update process, it must be guaranteed that the update does not lead to an infeasible vehicle state for the velocity-planning algorithm, e.g., when the vehicle is approaching a turn already utilizing full tire forces under braking, and the acceleration limits are suddenly decreased in front of the vehicle. Therefore, the updates only take place outside the planning horizon of the algorithm.

The difference between subsequently obtained acceleration limits $\bar{a}_\mathrm{x/y}(s_m)$ at identical local path coordinates $s_m$ can be controlled via the maximum change between the stored values $\Sigma_{\bar{a}}(s_\mathrm{glo})$. The slope of the interpolated values $\tilde{\Sigma}_{\bar{a}}(s_\mathrm{glo})$ can be used to calculate the maximum error when applying a particular step size $\Delta s_m$ in path planning.

\subsection{Recursive feasibility}
\label{subsec:slackVariable}
The minimum-time optimization problem tends to produce solutions with many active constraints, as it maximizes the tire utilization. It follows from this property, that ensuring recursive feasibility is a highly relevant aspect for the application of such an algorithm, and should be achieved via the terminal constraints (\ref{eq:v_recursiveInfeas}) and (\ref{eq:acc_recursive}) by making a worst-case assumption about the curvature $\kappa(s_\mathrm{f})$ and acceleration limits $\bar{a}_\mathrm{x/y}(s_\mathrm{f})$ at the end of the optimization horizon. This property holds as long as the optimization problem is shifted by an integer multiple of the discretization $\Delta s_m$ while the relations of the local path coordinates $s_m$ with the curvature $\kappa(s_m)$ and the acceleration limits $\bar{a}_\mathrm{x}(s_m)$ and $\bar{a}_\mathrm{y}(s_m)$ remain constant. This cannot be ensured since the path planner \cite{Stahl2019} might slightly vary the target path due to an obstacle entering the planning horizon, or due to discretization effects. This leads to a deviation in the curvature profile $\kappa(s_m)$ and deviations in the admissible accelerations $\bar{a}_{x}(s_m)$ and $\bar{a}_{y}(s_m)$ since the local coordinate $s_m$ might refer to a different point in global coordinates $s_\mathrm{glo}$ now, see Fig. \ref{fig:map_discretization}.

To mitigate this deficiency, we introduce slack variables $\epsilon$ based on the exact penalty function approach \cite{Kerrigan2000}. This strategy ensures that the hard-constrained solution is recovered if it is feasible, and therefore the solution is not altered by addition of the slacks unless it is mandatory. The nature of the combined acceleration constraint (\ref{eq:frictionConstraint}) allows for a straightforward interpretation of the slack variables as a violation of $\epsilon$ in \SI{}{\percent}. Together with the upper bound on the slack variables in (\ref{eq:EpsilonConstraint}), we can therefore state that the optimization problem is always feasible as long as the maximum required violation is limited to $\epsilon_{\mathrm{max}}$. In case no solution is found within the specified tolerance band, a dedicated failure-handling strategy is employed within the trajectory planning framework. We wish to point out that a suitable scaling of the slack variables is crucial to achieve sufficiently tight tolerances $\varepsilon_\mathrm{QP,tol}$ when using a numerical \gls{QP} solver. We therefore employ a variable transformation with $\epsilon = \zeta\epsilon_n$ and optimize over $\epsilon_n$ instead. Realistic values for the maximum slack variable $\epsilon_\mathrm{max}$ were found to be around \SI{3}{\percent} in extensive simulations on different race tracks (Berlin (Germany), Hong Kong (China), Indianapolis Motor Speedway (USA), Las Vegas Motor Speedway (USA), Millbrook (UK), Modena (Italy), Monteblanco (Spain), Paris (France), Upper Heyford (UK), Zalazone (Hungary)) and obstacle scenarios. We consider this to be an acceptable tolerance level and believe it will be difficult to achieve significantly tighter guarantees in the face of the scenario complexity we tackle in \cite{Stahl2019}.

\section{Results}
In this section, the results achieved with the presented velocity \gls{mpSQP} will be presented. We conducted the experiments on our Hardware-in-the-Loop simulator, which consists of a Speedgoat Performance real-time target machine, where validated physics models of the real race car in combination with realistic sensor noise are implemented. An additional NVIDIA Drive PX2 receives this sensor feedback and calculates the local trajectories. A Speedgoat Mobile real-time target machine transforms this trajectory input into low level vehicle commands to close the loop to the physics simulation. Therefore, we used the DevBot 2.0 data: $m_\mathrm{v} = \SI{1160}{\kilogram}$, $P_\mathrm{max} = \SI{270}{\kilo\watt}$, $F_\mathrm{max} = \SI{7.1}{\kilo\newton}$,  $F_\mathrm{min} = \SI{-20}{\kilo\newton}$, $c_\mathrm{r} = \SI{0.85}{\kilogram\per\meter}$. The results in this section have been produced with the velocity planner parametrizations given in Table \ref{tab:keyfacts_SQP}.
\begin{table}[!tb]
	\renewcommand{\arraystretch}{1.1}
	\caption{Emergency- and Performance-SQP parametrization.}
	\label{tab:keyfacts_SQP}
	\centering
	\begin{tabular}{|c c c c c c|}
		\hline
		 & Parameter & Unit & \multicolumn{2}{c}{Value} & \\
		 & & & Performance & Emergency & \\
		\hline
		 & $M$ & - & 115 & 50 & \\
		 & $N$ & - & 12 & 5 & \\
		 & $\delta_\mathrm{a}$ & \SI{}{\meter\per\second\squared} & 0.1 & inactive & \\
		 & $\epsilon_\mathrm{max}$ & \SI{}{\percent} & 3.0 & 3.0 & \\
		 & $\rho_{\mathrm{j}}$ & - & $3e^2$ & 0.0 & \\
		 & $\rho_{\mathrm{\epsilon,l}}$ & \SI{}{\meter\squared\per\second\squared} & $1e^5$ & $5e^4$ & \\
		 & $\rho_{\mathrm{\epsilon,q}}$ & \SI{}{\meter\squared\per\second\squared} & $1e^4$ & $1e^3$ & \\
		 & $n_{\mathrm{SQP,max}}$ & - & 20 & 20 & \\
		 & $\Delta t_{\mathrm{max}}$ & \SI{}{\milli\second} & $300$ & $100$ & \\
		 & $\beta$ & - & $0.5$ & $0.5$ & \\
		 & $\bar{\varepsilon}_\mathrm{SQP,tol}$ & - & $1e^0$ & $1.5e^0$ & \\
		 & $\hat{\varepsilon}_\mathrm{SQP,tol}$ & - & $1e^0$ & $1.5e^0$ & \\
		 & $\varepsilon_\mathrm{QP,tol}$ & - & $1e^{-2}$ & $1e^{-2}$ & \\
		\hline
	\end{tabular}
\end{table}

We show results for two types of offered paths: performance and emergency. The emergency path is identical to the performance one, except for a coarser spatial discretization, and the fact that the velocity planner tries to stop as soon as possible on the emergency line. The optimization of the emergency line requires therefore fewer variables $M + N$. This formulation reduces the necessary calculation time of the emergency line that must be updated more frequently for safety reasons.

\subsection{Objective function design}
\begin{figure}[!tb]
	\centering
	\input{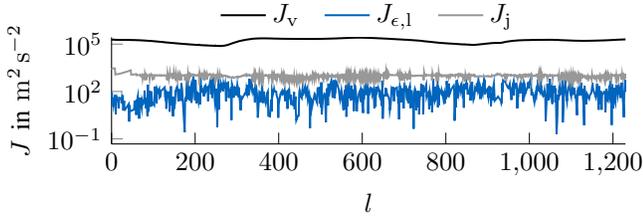}
	\caption{Cost terms of the objective function $J$ being minimized within the performance velocity profile. $J_\mathrm{v}$ shows the most significant influence on the solution due to its value range compared to the other objective terms $J_\mathrm{j}$, $J_\mathrm{\epsilon,l}$ and $J_\mathrm{\epsilon, q}$ (not displayed as it equals almost 0). Symbol $l$ denotes the number of the optimized velocity profiles.}
	\label{fig:costPerf}
\end{figure}

To explain the chosen values of the penalty weights $\rho$, we show the values of the single objective function terms in $J(\boldsymbol{z})$ being minimized during the calculation of the performance velocity profile in Fig. \ref{fig:costPerf}. The symbol $l$ denotes the number of the optimized velocity profiles during the driven lap (including the race start) as well as the vehicle's stopping scenario. It can clearly be seen that the velocity term $J_\mathrm{v}$ has the highest relative impact on the optimum solution. Its value range is at least two orders of magnitude higher than the slack penalty terms $J_\mathrm{\epsilon,l}$, $J_\mathrm{\epsilon,q}$ (not displayed) and the jerk penalty $J_\mathrm{j}$. The penalty weight $\rho_\mathrm{\epsilon,l}$ on the linear slack term  was chosen to increase the value of $J_\mathrm{\epsilon,l}$ to be one order of magnitude higher than $J_\mathrm{v}$ if $\epsilon_\mathrm{max}$ was fully exploited on all the slack variables $\epsilon_n$. Therefore, $J_\mathrm{\epsilon,l}$ prevents the solver from permanent usage of tire slack $\epsilon$ for further lap time gains. As $\rho_\mathrm{\epsilon,q}$ is applied to the squared values $\epsilon_n$, it is sufficient to keep the magnitude of $\rho_\mathrm{\epsilon,q}$ one order smaller than $\rho_\mathrm{\epsilon,l}$. To provide a smooth velocity profile, we set the jerk penalty $\rho_\mathrm{j}$ to increase the value of $J_\mathrm{j}$ to be higher than the slack penalty terms during normal operation. By this, effects on a possible lap time loss stay as small as possible, whilst a smoothing effect in the range of numerical oscillations on the velocity and acceleration profile is still visible.

We further integrated a calculation time limit $\Delta t_\mathrm{max}$ for the velocity optimization and a maximum \gls{SQP} iteration number $n_\mathrm{SQP,max}$. In case of reached limits, the algorithm would return the last suboptimal but driveable solution. The \gls{SQP} never reached these limits during our experiments, and they can be considered as safety limitations. Instead, the \gls{SQP}-algorithm always terminated due to the reached tolerance criteria $\bar{\varepsilon}_\mathrm{SQP,tol}$ and $\hat{\varepsilon}_\mathrm{SQP,tol}$, cf. (\ref{eq:eps_sqp_bar}) and (\ref{eq:eps_sqp_hat}).

\subsection{Energy Strategy}
\begin{figure}[!tb]
	\centering
	\input{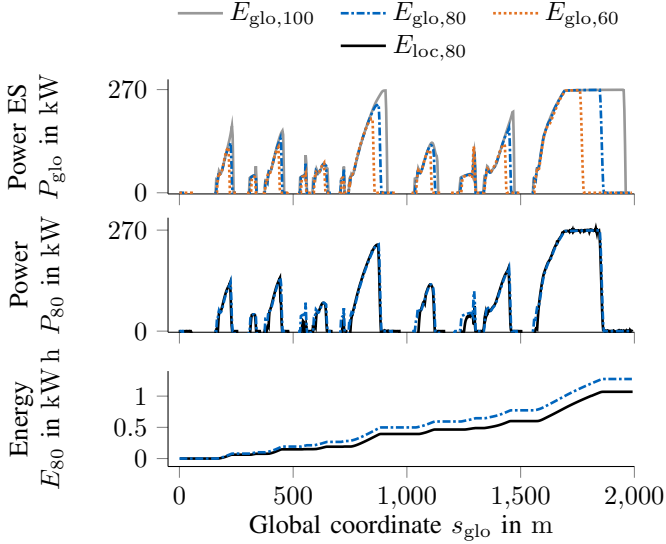}
	\caption{Top: Optimal power usage $P_\mathrm{glo} (s_\mathrm{glo})$ from the \gls{ES} that should be requested whilst driving on the Modena (Italy) race track. Mid: Power planned locally $P_\mathrm{loc,80}(s_\mathrm{glo})$ compared to $P_\mathrm{glo,80} (s_\mathrm{glo})$ resulting in an absolute drift in the energy demand between $E_\mathrm{glo,80}(s_\mathrm{glo})$ and $E_\mathrm{loc,80}(s_\mathrm{glo})$ of approx. \SI{15.8}{\percent} within one race lap (bottom).}
	\label{fig:optimalPowerUsage}
\end{figure}

As stated in Subsection \ref{subsec:efficient_implementation}, the presented \gls{mpSQP} is able to implement our global race \gls{ES} \cite{Herrmann2020, Herrmann2019}. The \gls{ES} is pre-computed offline and re-calculated online due to disturbances, unforeseen events during the race and model uncertainties. Through this, we account for the limited amount of stored battery energy and the thermodynamic limitations of the electric powertrain. Therefore, the \gls{ES} delivers the maximum permissible power $P_\mathrm{max} (s_\mathrm{glo})$ in order to reach the minimum race time, see Fig. \ref{fig:architecture}. It takes the following effects into account: 
\begin{itemize}
	\item the vehicle dynamics in the form of an \gls{NDTM} including a nonlinear tire model;
	\item the electric behavior of battery, power inverters and electric machines, i.e., the power losses of these components during operation;
	\item the thermodynamics within the powertrain transforming power losses into temperature contribution.
\end{itemize}
Fig. \ref{fig:optimalPowerUsage} depicts the output of the \gls{ES} computed offline (top). We varied the amount of energy available for one race lap by the three values \SI{100}{\percent} ($E_\mathrm{glo,100}(s_\mathrm{glo})$), \SI{80}{\percent} ($E_\mathrm{glo,80}(s_\mathrm{glo})$) and \SI{60}{\percent} ($E_\mathrm{glo,60}(s_\mathrm{glo})$). The optimal power usage $P_\mathrm{glo}(s_\mathrm{glo})$ belonging to these energy values $E_\mathrm{glo}(s_\mathrm{glo})$ is depicted in the first diagram. The positive values in $P_\mathrm{glo}(s_\mathrm{glo})$ become the parametric input of the power constraint within the velocity optimization (\ref{eq:EnergyConstraint}). By this formulation, we only restrict the vehicle's acceleration power but leave the braking force unaffected. This experiment consists of one race lap with constant maximum acceleration values $\bar{a}_\mathrm{x/y}$ on the Modena (Italy) race circuit.

The power usage $P_\mathrm{loc,80}(s_\mathrm{glo})$ locally planned by the \gls{mpSQP} is shown in the middle plot of Fig. \ref{fig:optimalPowerUsage}. The positive power values in $P_\mathrm{loc,80}(s_\mathrm{glo})$ remained below the maximum power request allowed by the global strategy $P_\mathrm{glo,80}(s_\mathrm{glo})$. Differences between the globally optimal power usage $P_\mathrm{glo}(s_\mathrm{glo})$ and the locally transformed power $P_\mathrm{loc}(s_\mathrm{glo})$ stem from different model equations in both - offline \gls{ES} and online \gls{mpSQP} - optimization algorithms. The velocity planner's point mass is more limited in its combined acceleration potential due to the diamond-shaped acceleration constraint (\ref{eq:frictionConstraint}) compared to the vehicle dynamics model (\gls{NDTM}) in the \gls{ES}. The \gls{NDTM} overshoots the dynamical capability of the car slightly in edge cases due to parameter-tuning difficulties. Furthermore, the effect of longitudinal wheel-load transfer is considered within the \gls{NDTM}. For these reasons, the point mass model accelerates less but meets the maximum admissible power $P_\mathrm{glo}(s_\mathrm{glo})$ on the straights.
The accumulated error in this experiment can be expressed by the energy demand $E_\mathrm{loc}$ resulting from
\begin{equation}
	E_\mathrm{loc} = \int{F_\mathrm{x,p}(s_\mathrm{glo}) \mathrm{d}s_\mathrm{glo}}.
\end{equation}
In total, an energy amount of $E_\mathrm{glo} = \SI{1.27}{\kilo\watt\hour}$ was allowed whereas $E_\mathrm{loc} = \SI{1.07}{\kilo\watt\hour}$ was used, implying \SI{15.8}{\percent} drift. With the help of a re-calculation strategy adjusting the \gls{ES} during the race, this error can be significantly reduced. This feature was switched off during this experiment to isolate the working principle of the race \gls{ES}, i.e., the interaction between global and local planners.

The effect of the \gls{ES} on the vehicle speed $\boldsymbol{v}_\mathrm{P80}$ is depicted in Fig. \ref{fig:ES_local}. The left part (A) of both plots consists of a scenario where the race car is accelerating with the maximum available machine power of $P_\mathrm{max} = \SI{270}{\kilo\watt}$ and a subsequent straight where the \gls{ES} forces the vehicle to coast ($P_\mathrm{glo,80}(s_m) = P_\mathrm{loc,80}(s_m) = \SI{0}{\kilo\watt}$). Acceleration without any power restriction on this straight would have resulted in the velocity curve $\boldsymbol{v}_\mathrm{P100}$ meaning a slightly higher top speed by approx. $\SI{4}{\percent}$ or $\SI{8}{\kilo\meter\per\hour}$. The second part (B) of the shown planning horizon ensures recursive feasibility. Here, we force the velocity variable $v(s_N)$ to reach $v_\mathrm{end}$ (\ref{eq:vendConstraint}).
\begin{figure}[!tb]
	\centering
	\input{./ressources/energy_strategy/EMS_lift-cost_ModenaE80.tex}
	\caption{The effect of the \gls{ES} on one solution of the velocity-planning problem on the vehicle speed $\boldsymbol{v}_\mathrm{P80}$. In part (A), the race car is accelerating with the maximum available machine power of $P_\mathrm{\Sigma} = \SI{270}{\kilo\watt}$ and is forced to coast ($P_\mathrm{glo}(s_m) = P_\mathrm{loc}(s_m) = \SI{0}{\kilo\watt}$) thereafter. Acceleration without any power restriction is denoted by $\boldsymbol{v}_\mathrm{P100}$. Part (B) of the planning horizon ensures recursive feasibility.}
	\label{fig:ES_local}
\end{figure}

\subsection{Variable acceleration limits}
\label{subsec:VarFriction}
\begin{figure}[!tb]
	\centering
	\input{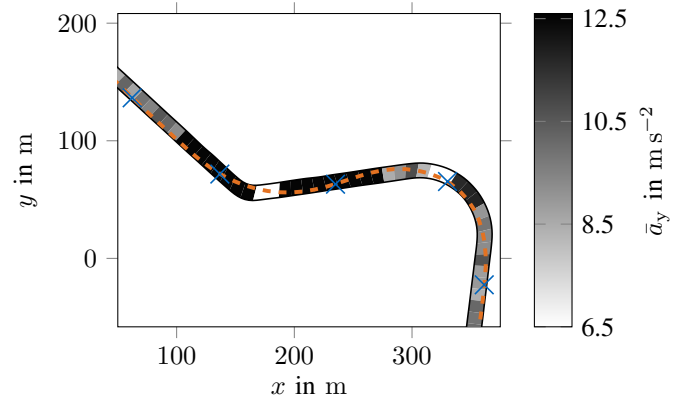}
	\caption{Race track map displaying the spatially variable acceleration potential $\Sigma_{\bar{a}}(s_\mathrm{glo})$ including the driven path and markers starting from the global $s$-coordinate of \SI{1400}{\meter} at \SI{100}{\meter} gaps.}
	\label{fig:tpamap}
\end{figure}

\begin{figure}[!tb]
	\centering
	\input{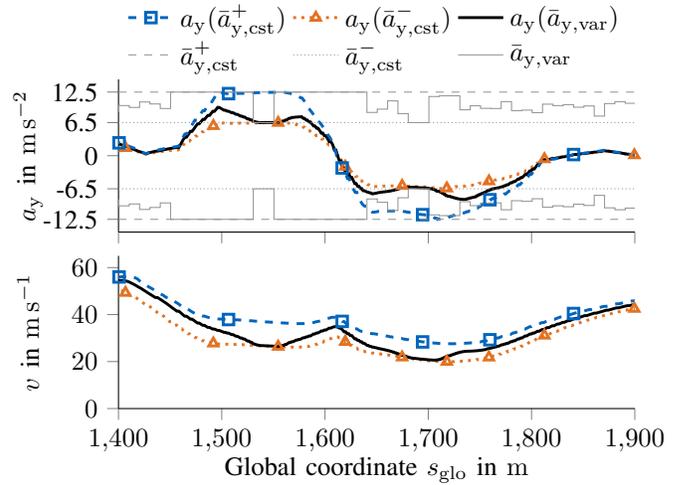}
	\caption{Comparison of velocity $v(s_\mathrm{glo})$ and lateral acceleration $a_\mathrm{y}(s_\mathrm{glo})$ profiles resulting from spatially and temporally varying acceleration coefficients $\bar{a}_\mathrm{y}(s_\mathrm{glo})$.}
	\label{fig:velocityComp_varFriction}
\end{figure}

Fig. \ref{fig:tpamap} shows a locally variable acceleration map to determine the values of $\bar{a}_\mathrm{y} (s)$ along the driven path.
We conducted three experiments with a constant acceleration potential of  $\bar{a}_\mathrm{x}$ = \SI{12.5}{\meter\per\second\squared} and varied $\bar{a}_\mathrm{y}$ in three different ways:
\begin{itemize}
	\item $\bar{a}_\mathrm{y,cst}^+$ has a constant value of \SI{12.5}{\meter\per\second\squared} along the entire track, describing a high friction potential.
	\item $\bar{a}_\mathrm{y,cst}^-$ has a constant value of \SI{6.5}{\meter\per\second\squared} along the entire track, describing a low friction potential.
	\item $\bar{a}_\mathrm{y,var}(s)$ has a variable value in the range of \SIrange{6.5}{12.5}{\meter\per\second\squared} with the steepest gradients between $s_\mathrm{glo}$ = \SI{1520}{} - \SI{1550}{\meter} and $s_\mathrm{glo}$ = \SI{1670}{} - \SI{1700}{\meter}, as shown in Fig. \ref{fig:tpamap}.
\end{itemize}
The results of these experiments can be seen in Fig. \ref{fig:velocityComp_varFriction}. We depicted the acceleration potentials $\bar{a}_\mathrm{y}$ of the three scenarios in combination with the planned lateral acceleration $a_\mathrm{y}(s_\mathrm{glo})$ (top) and the vehicle velocity $v(s_\mathrm{glo})$ (bottom).
In both scenarios with constant $\bar{a}_\mathrm{y,cst}$ values, the planner leverages the entire admissible lateral potentials which can be seen between $s_\mathrm{glo}$ = \SI{1500}{} - \SI{1600}{\meter} and $s_\mathrm{glo}$ = \SI{1650}{} - \SI{1700}{\meter}. The results achieved using the variable $\bar{a}_\mathrm{y,var}(s)$ are interesting: The planned lateral acceleration $a_\mathrm{y}(s)$ stayed within the boundaries of the low- and high-friction experiments. The planner handles the drop of $\bar{a}_\mathrm{y,var}(s)$ of \SI{50}{\percent} at $s_\mathrm{glo}$ = \SI{1520}{\meter} by reducing $a_\mathrm{y}(s)$ in advance, while still fully leveraging $\bar{a}_\mathrm{y,var}(s)$. This results in a vehicle velocity of $v(s_\mathrm{glo})$ in the low-friction scenario. The same is true for $s_\mathrm{glo}$ = \SI{1670}{\meter}. Also during the rest of the shown experiment, the oscillating values of $\bar{a}_\mathrm{y,var}(s)$ can be handled by the \gls{mpSQP} algorithm. This behavior allows us to fully exploit the dynamical limits of the race car on a track with variable acceleration potential.

\begin{figure}[!tb]
	\centering
\begin{tikzpicture}

\begin{axis}[
width=0.91\columnwidth,
legend cell align={left},
legend style={draw=none, fill opacity=0.7, text opacity=1, at={(0.95,0.05)}, anchor=south east},
tick align=outside,
tick pos=both,
xlabel={\(\displaystyle a_\mathrm{y}\) in \(\displaystyle \SI{}{\meter\per\second\squared}\)},
xmin=-13, xmax=13,
xtick style={color=black},
ylabel={\(\displaystyle a_\mathrm{x}\) in \(\displaystyle \SI{}{\meter\per\second\squared}\)},
ymin=-13, ymax=13,
ytick style={color=black},
axis equal
]
\addplot [semithick, TUMgray]
table {%
12.5 0
0 12.5
-12.5 0
0 -12.5
12.5 0
};
\addlegendentry{$\bar{\mu}$}
\addplot [semithick, TUMgray, dashed]
table {%
12.875 0
0 12.875
-12.875 0
0 -12.875
12.875 0
};
\addlegendentry{$\bar{\mu} + \epsilon_\mathrm{max}$}
\addplot [semithick, black, mark=x, mark size=2, mark options={solid}, only marks]
table {%
6.36540115861395 0.259979824228829
6.37600716257157 0.239331836268919
6.38838353151047 0.215288501144083
6.39365551423362 0.17319069033678
6.39983687036558 -0.191247141379667
6.39427178355802 -0.203682550827989
6.39051246249619 -0.211260746597016
6.3980080765557 -0.196865205656963
6.4084380071172 -0.17674776135328
6.42186824602321 -0.150882276808416
6.43836173871728 -0.120191032306971
6.44460421788364 -0.108361908637611
6.45401753722397 -0.0905388115384371
6.4666923961484 -0.0664710895405998
6.48272471278398 -0.0359103118748364
6.47716963778308 -0.0467878641121751
6.4738495110805 -0.0532686613146198
6.47283749112032 -0.0552079529179352
6.47420322971285 -0.0524754119338421
6.47509629064549 -0.0505510017142053
6.47936174222969 -0.0406451780201206
6.48719753083747 -0.0252508790112038
6.49869032346086 -0.000205846350273209
6.45146428647928 0.0927034512580084
6.40969506283826 0.173223402894361
6.37290247075512 0.244154777483045
6.34072450043628 0.306202012848193
6.21409003697395 0.54988511059897
6.10519221770569 0.759448387450706
6.01233383795186 0.93815874778505
5.93408169131434 1.08973634611999
5.57549306775433 1.77942641258004
5.23230722490707 2.43947506426086
4.89999122565531 3.07859929215438
4.57418542496444 3.705182559161
4.25064395869007 4.32738935849508
3.92516571150602 4.95329808756079
3.59351745946815 5.59104675415754
3.25134973514008 6.12146128876168
2.89187448534579 6.12135220804524
2.50311794285366 6.12123399320268
2.08518749611413 6.1211225035194
1.63792586261007 4.59257400668763
0.849949129007392 0.583538538021089
0.0264082708329335 -4.24511112485084
-0.776578208686932 -8.16353573726837
-1.50361158339827 -9.60840000521127
-2.13590771690982 -8.3925083847949
-2.70157526558528 -7.30473447447353
-3.21394382987101 -6.31946332023177
-3.68422072148747 -5.41555835463881
-4.12188240033254 -4.57395896465101
-4.53502299912809 -3.77951533627293
-4.93056134637437 -3.01892192491321
-5.31441994095701 -2.28081932802802
-5.49162757843712 -1.94010629966778
-5.68116724365608 -1.57567838968888
-5.88553080872553 -1.18274363953551
-6.10749153158937 -0.755967318543309
-6.0750718058806 -0.818386917906976
-6.04190231803223 -0.882884502868491
-6.00761592842607 -0.948888159175792
-5.97184790955637 -1.01773241473206
-5.96064454266478 -1.03933064270413
-5.9516934092523 -1.05659058745556
-5.94467408398098 -1.0701284415773
-5.93921325552876 -1.08066197345042
-5.92139389558678 -1.11495494037598
-5.90493673991039 -1.14662155086651
-5.8893722068355 -1.1765648169179
-5.87415228988182 -1.20594468259397
-5.89761099097152 -1.1608298461089
-5.92620718685209 -1.10582813879192
-5.95965600581812 -1.04148763905102
-5.99762483544355 -0.968448846452222
-6.01493063457122 -0.93514002750033
-6.03871763526318 -0.889362160442897
-6.06882898044052 -0.83141725560963
-6.10507345683622 -0.76167541765975
-6.10613417675905 -0.759589869508574
-6.11282035706296 -0.746890740911275
-6.12492389683448 -0.723568455486294
-6.14220020293302 -0.690301751672781
-6.20830143111586 -0.563143382748658
-6.28971499049012 -0.406541143721205
-6.38745052930192 -0.218566883088566
-6.50270540338913 -0.000334389278619673
-6.45541131052754 0.0873488735282942
-6.42361815103718 0.148462317635289
-6.40547432096662 0.183335527391183
-6.39911209376683 0.194289055115346
-6.29204324532918 0.400176842488039
-6.20620996093511 0.565234088798678
-6.13918608530241 0.694121164639715
-6.08897661592865 0.790674170553871
-6.04037430835605 0.884138304401111
-6.00281884164628 0.956449010559726
-5.97557677330077 -0.0511098818548448
-5.88900850631569 -1.1748411551665
-5.68290976907365 -1.5711083179557
-5.46069007386102 -1.99813481329517
-5.22252702866976 -2.45608338587434
-4.96836068360284 -2.94479125959783
-4.59980035747416 -3.65346650714982
-4.20857850019135 -4.40569328343554
-3.7978506368443 -5.19539681678379
-3.37077093852549 -6.0165119974382
-2.92075633834128 -6.88173319078044
-2.45865627146164 -7.77037793274239
-1.98923794806436 -8.67416513380143
-1.51723917056124 -9.58841091666203
-1.05973017900771 -10.4693496873452
-0.597593340732667 -7.40995063907531
-0.286942366131057 -2.71836621292297
};
\addlegendentry{$\mu_\mathrm{o}(\bar{\mu}_\mathrm{l})$}
\addplot [semithick, TUMblue, mark=x, mark size=2, mark options={solid}, only marks]
table {%
	12.2071583517336 0.292527300926229
	12.2195891055405 0.280140709081432
	12.2364293112047 0.263310257858101
	12.2407593095637 0.248432733033529
	12.2493657935697 -0.158472619391251
	12.2468019007223 -0.253765078273161
	12.2449352377964 -0.255856415475189
	12.2651671828197 -0.235677577727947
	12.2907810434116 -0.210105156235446
	12.3216826774763 -0.179230342967325
	12.3577599048789 -0.143459674596959
	12.3732675175378 -0.127952870320641
	12.3945955972486 -0.106618711946296
	12.4216862959093 -0.0795173579599224
	12.4544635011358 -0.046728947618986
	12.4447083126182 -0.0564723062772481
	12.439710474133 -0.0614609263837894
	12.4394618541367 -0.0617039825800419
	12.4439347750598 -0.0572316545205321
	12.4474699329468 -0.0536968432867575
	12.4574898461305 -0.0424992796991694
	12.4740546342419 -0.0259548311817772
	12.497109357959 -0.00036235370257065
	12.4062941656129 0.0919869026977572
	12.3223627765822 0.175882343432549
	12.2450826301614 0.253149536512908
	12.174284265085 0.323960316451859
	11.920394822388 0.577890947505738
	11.6924210310633 0.8059367292945
	11.4890847357985 1.00937588074794
	11.3092500628608 1.18995989699258
	10.5900431358357 1.90928009627396
	9.88300650958915 2.6164264214365
	9.18626966736541 3.31326113524349
	8.4972868370419 4.00232376208395
	7.81292861420571 4.68674060669236
	7.12955158544455 5.37015633566504
	6.4430461147305 6.05668515120524
	5.74886323492163 6.12057372616309
	5.03099745124098 6.12058324952878
	4.28830088999046 6.08325009355115
	3.52020687781077 4.6250445573386
	2.71368009451386 0.533869418769366
	1.37580980598694 -4.49117695077465
	0.0419740927200973 -8.75198922089943
	-1.2257092088105 -10.9505549512234
	-2.39011718348157 -10.1106544393668
	-3.46328249301565 -9.03760147697679
	-4.46114232347675 -8.03975655613965
	-5.39618565442041 -7.10481770058181
	-6.27898568017285 -6.22284965878694
	-7.11828186566953 -5.38356357787029
	-7.92109560488306 -4.58074208191696
	-8.69267861451919 -3.80926487820948
	-9.43649004816579 -3.06537614759602
	-9.7925176799688 -2.70936815473066
	-10.1637304823182 -2.33815806976633
	-10.5517585472076 -1.95006447782778
	-10.9580472560711 -1.54363204150583
	-10.8906356221176 -1.61090823297137
	-10.8240434066279 -1.67617995291663
	-10.7579113542295 -1.74209093706863
	-10.6917121265867 -1.80810393669004
	-10.6723630841252 -1.82725143647662
	-10.6580587334914 -1.84136512765722
	-10.6481160248453 -1.85112953320299
	-10.6417764046726 -1.85730071140045
	-10.6140305035499 -1.88487691795806
	-10.5906124259885 -1.90811530297586
	-10.570724091451 -1.92779637988764
	-10.5534771759427 -1.94377088827929
	-10.608061240415 -1.88901315996565
	-10.6708203118203 -1.82617180559343
	-10.7406245192607 -1.75638234958601
	-10.8161932021637 -1.68092422142947
	-10.8513448899336 -1.64597664008198
	-10.8967689355615 -1.60083620129768
	-10.9513533407191 -1.54659374828972
	-11.0138275029019 -1.48449651897083
	-11.0087022471476 -1.49001525897085
	-11.0129743179259 -1.48872555351917
	-11.025418646747 -1.47661563638847
	-11.0449671251288 -1.45734190957845
	-11.1492619077862 -1.3532577279127
	-11.2718861541101 -1.23078432963507
	-11.411589930261 -1.09117912724434
	-11.566861033154 -0.935964270104386
	-11.4154833899228 -1.08735944161148
	-11.2702049498458 -1.23262790548497
	-11.1314213310844 -1.37137747757755
	-10.9992874330962 -1.50314715298059
	-10.6850530992237 -1.81731430745597
	-10.364962423285 -2.13732626293734
	-10.0408603330974 -2.46133511001827
	-9.71422066147931 -2.78786184294699
	-9.36360789614376 -3.13834557155207
	-9.00518569237658 -3.49662142951171
	-8.63983683792547 -3.86180932781952
	-8.26809156962371 -4.23338260590076
	-7.81345173655837 -4.68784205564268
	-7.34922649575972 -5.15065634807683
	-6.87690943010699 -5.62281350520523
	-6.39742489295026 -6.10216351873198
	-5.78848781780712 -6.71099079700267
	-5.17785295299781 -7.32154347524082
	-4.5701434764068 -7.92920493362822
	-3.96935975368043 -8.52999497588314
	-3.36800621280888 -9.13146706757248
	-2.77861916170379 -9.72124453167724
	-2.20570390226206 -10.2952188906393
	-1.6532034391045 -10.8486806038374
	-1.13798606995008 -11.3618605574441
	-0.636927547857617 -8.38620834834904
	-0.28693933887012 -2.71830051403027
};
\addlegendentry{$\mu_\mathrm{o}(\bar{\mu}_\mathrm{h})$}
\addplot [semithick, white!50.19607843137255!black, forget plot]
table {%
6.5 0
0 12.5
-6.5 0
0 -12.5
6.5 0
};
\addplot [semithick, white!50.19607843137255!black, dashed, forget plot]
table {%
	6.695 0
	0 12.875
	-6.695 0
	0 -12.875
	6.695 0
};
\addplot [semithick, TUMred, dashed]
table {%
-12.5 6.12068965517241
12.5 6.12068965517241
};
\addlegendentry{$\bar{a}_\mathrm{x,m}$}
\end{axis}

\end{tikzpicture}
	\caption{Acceleration limit model including operating points at low- and high-friction experiments ($\bar{\mu}_\mathrm{l}$ and $\bar{\mu}_\mathrm{h}$) for a planning horizon ranging from the global coordinate of $s_\mathrm{glo}$ = \SIrange{1500}{1800}{\meter}.}
	\label{fig:friction_diamond}
\end{figure}
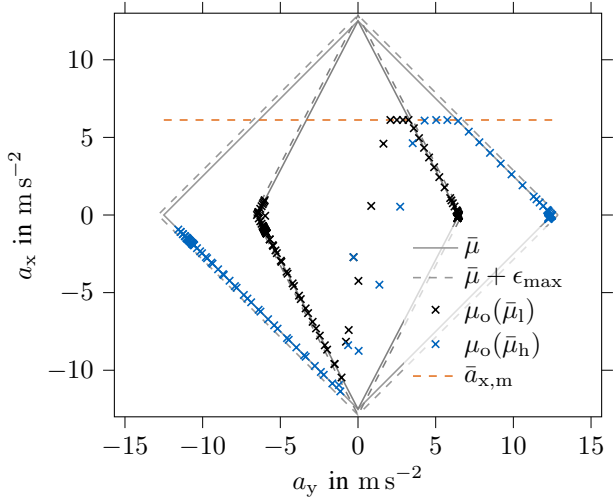

A further indicator that the velocity planner utilizes the full acceleration potential is shown in Fig. \ref{fig:friction_diamond}. Here, we depicted the vehicle's optimized operating points $\mu_\mathrm{o}(s)$ regarding combined acceleration $a_\mathrm{x}(s)$ and $a_\mathrm{y}(s)$ within the planning horizon of $s_\mathrm{glo}$ = \SI{1500}{} - \SI{1800}{\meter}. Both solid diamond shapes express the given constant acceleration limits of the high-friction scenario, $\bar{\mu}_\mathrm{h} = \bar{a}_\mathrm{x}$ \& $\bar{a}_\mathrm{y,cst}^+$ = \SI{12.5}{} \& \SI{12.5}{\meter\per\second\squared}, and the low-friction scenario $\bar{\mu}_\mathrm{l} = \bar{a}_\mathrm{x}$ \& $\bar{a}_\mathrm{y,cst}^-$ = \SI{12.5}{} \& \SI{6.5}{\meter\per\second\squared}. The horizontal red dashed line indicates the maximum available electric machine acceleration $\bar{a}_\mathrm{x,m} = \frac{F_\mathrm{max}}{m_\mathrm{v}}$. From this diagram we see that the vehicle operates at the limits of the given acceleration constraints in both scenarios. This means the \gls{mpSQP} leverages the maximum acceleration potential in combination with fully available cornering potential.

\subsection{Solver comparison}
The number of variables in the performance profile within one \gls{QP} for the performance trajectory is \SI{126}{}, including \SI{810}{} constraints. From the problem formulation (\ref{eq:discreteObjective}), a small number of non-zero entries in the matrices $\boldsymbol{A}$ and $\boldsymbol{P}$ of \SI{2295}{} in total arises with constant entries in the problem's Hessian $\boldsymbol{P}$.

\begin{figure}[!tb]
	\centering
	\input{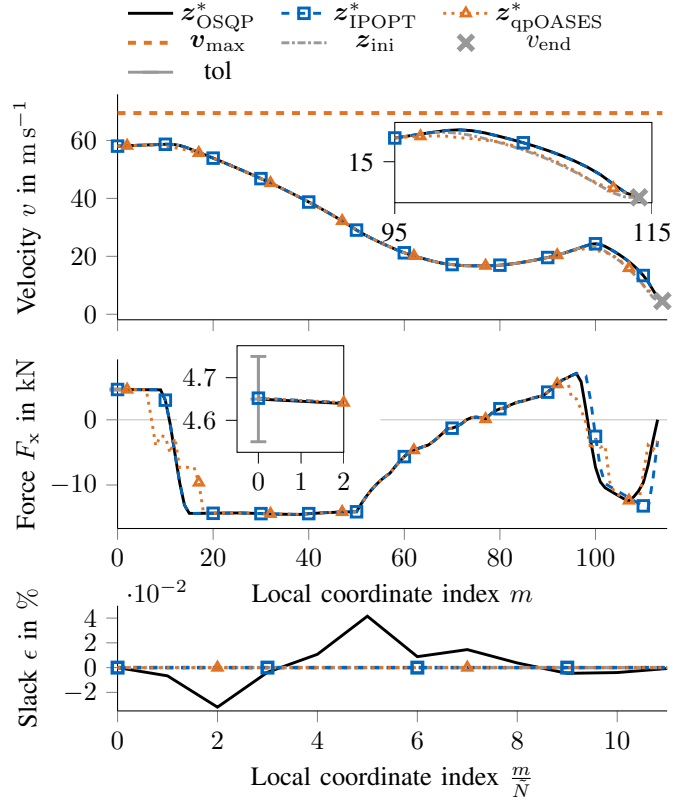}
	\caption{Comparison of the solution of the \gls{mpSQP} (internal \gls{QP}s solved by the \gls{ADMM} solver \gls{OSQP} \cite{Stellato2020}) with the general \gls{NLP} interior point solver \gls{IPOPT} \cite{Wachter2006} interfaced by CasADi \cite{Andersson2019} and the active set solver \gls{qpOASES} \cite{Ferreau2014}. The chosen scenario includes the vehicle heading towards a narrow right-hand bend at a high velocity of almost \SI{200}{\kilo\meter\per\hour}. The optimization horizon spans the curve, and includes the subsequent straight.}
	\label{fig:solution_SQPvsNLP}
\end{figure}

Fig. \ref{fig:solution_SQPvsNLP} contains a comparison
of the \gls{ADMM} solver \gls{OSQP} \cite{Stellato2020} and the active set solver \gls{qpOASES} \cite{Ferreau2014} as \gls{QP} solvers for an \gls{mpSQP} method as proposed in this paper. Furthermore, we solve the original \gls{NLP} using the interior point solver \gls{IPOPT} \cite{Wachter2006} interfaced via CasADi \cite{Andersson2019} to obtain a measure of solution quality for the proposed \gls{mpSQP}. Note that \gls{IPOPT} is widely used to benchmark the solution quality even if it is not specifically designed for embedded optimization. We chose a scenario where the vehicle is heading towards a narrow right-hand turn at a high velocity of almost \SI{200}{\kilo\meter\per\hour}. The optimization horizon spans this turn including the consecutive straight where positive acceleration occurs. Therefore, the solvers have to deal with high gradients for the longitudinal force $F_\mathrm{x}(s_m)$ and lateral acceleration $a_\mathrm{y}(s_m)$ stemming from curve entry and exit. The velocity plot shows that the optimal solution $\boldsymbol{z}^*$ almost equals the initial guess $\boldsymbol{z}_\mathrm{ini}$ for both the velocity vector $\boldsymbol{v}$ and the slack values $\boldsymbol{\epsilon}$. This behavior is expected, as the previous \gls{SQP} solution $\boldsymbol{z}^{l-1}(s_m)$ is shifted by the traveled distance and used as initialization $\boldsymbol{z}_\mathrm{ini}$. The optimization outputs $\boldsymbol{z}_\mathrm{OSQP}^*$ and $\boldsymbol{z}_\mathrm{IPOPT}^*$ overlap except at the end of the planning horizon where \gls{IPOPT} initially allows more positive longitudinal force $F_\mathrm{x}(s_m)$, resulting in more aggressive braking to fulfill the hard constraint $v_\mathrm{end}$. \gls{qpOASES}'s solution oscillates in the force $F_\mathrm{x}(s_m)$ at the steep gradients. All the algorithms keep the initially given longitudinal force $F_\mathrm{x}(s_\mathrm{s})$ within the specified tolerance of $\pm \SI{0.1}{\kilo\newton}$, with \gls{OSQP} matching the exact value (see magnified section in the second plot). The slack values $\boldsymbol{\epsilon}$ are close to zero. However, the \gls{OSQP}-solution shows small numerical oscillations within a negligible range of approx. $\pm \SI{0.04}{\percent}$. Nevertheless this behavior is typical for an \gls{ADMM} algorithm, and is thus noteworthy.

\begin{figure}[!tb]
	\centering
	\input{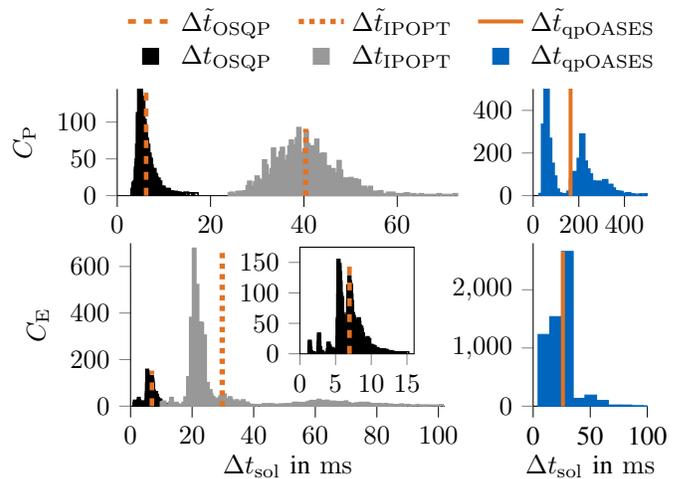}
	\caption{Solver runtimes for the \gls{mpSQP} with the local \gls{QP}s solved by \gls{OSQP} \cite{Stellato2020} or by the active set solver \gls{qpOASES} \cite{Ferreau2014}, compared to the problem solved by the general \gls{NLP} formulation passed to the interior point solver \gls{IPOPT} interfaced by CasADi \cite{Andersson2019}.}
	\label{fig:runtime_solvers}
\end{figure}

Apart from the solution qualities, we further analyzed the velocity optimization runtimes. The scenario consisted of two race laps, including a race start and coming to a standstill after the second lap on the Monteblanco (Spain) race circuit with a variable acceleration potential along the circuit, cf. Subsection \ref{subsec:VarFriction}. The histograms in Fig. \ref{fig:runtime_solvers} display the calculation times $\Delta t_\mathrm{sol}$ for the number of calls $C$ to optimize a speed profile. $C_\mathrm{P}$ and $C_\mathrm{E}$ denote the calls for the performance and the emergency lines, respectively. Their mean values are given by $\Delta \tilde{t}_\mathrm{sol}$. We wish to point out that the algorithm runtimes shown refer to an entire \gls{SQP} optimization process for a speed profile consisting of the solution of several \gls{QP}s in the case of the used solvers \gls{OSQP} or \gls{qpOASES}, which have been warm-started. The optimization runtimes on the specific CPUs are also summarized in Table \ref{tab:runtimes} where the ARM A57 is the NVIDIA Drive PX2 CPU. As we selected \gls{OSQP} for our application on the target hardware, we do not show additional solver times of \gls{IPOPT} or \gls{qpOASES} for the ARM A57 CPU.

Our \gls{mpSQP} in combination with the \gls{QP} solver \gls{OSQP} reaches nearly equal mean runtimes of \SIrange{6}{7}{\milli\second} for both velocity profiles on an Intel i7-7820HQ CPU and \SIrange{32}{34}{\milli\second} on an A57 ARM CPU. An amount of \SIrange{2}{5}{} \gls{SQP} iterations for the performance line was sufficient to reach the defined tolerances $\varepsilon_\mathrm{SQP}$. To optimize the emergency speed profile, a higher amount of \gls{SQP} iterations in the range of 5 - 10 was necessary. Therefore, the computational effort for the iterative linearizations increased on this profile. In contrast, it was possible to solve the single \gls{QP}s in less calculation time, as less than half the number of optimization variables $M + N$ in comparison to the performance profile are present. With maximum computation times of \SI{16.9}{\milli\second} (\SI{73.3}{\milli\second} A57 ARM) on the performance profile and \SI{15.0}{\milli\second} (\SI{77.1}{\milli\second} A57 ARM) for the emergency line, we managed to stay far below our predefined process-timeouts, see Table \ref{tab:keyfacts_SQP}.

The same optimization problem was formulated with the CasADi-language \cite{Andersson2019} as a general \gls{NLP} and passed to the interior point solver \gls{IPOPT}. The \gls{IPOPT} mean solver runtimes are at least approximately five times higher, with maxima of around \SI{0.1}{\second} on the emergency line being too long for vehicle operations at velocities beyond \SI{200}{\kilo\meter\per\hour}. On the emergency profile, the active set solver \gls{qpOASES} beats \gls{IPOPT} slightly in terms of its mean runtime $\Delta \tilde{t}_\mathrm{sol}$ but consumes four times the \gls{IPOPT} computation time to generate the performance speed profile, see Table \ref{tab:runtimes}. This behavior is rational, as a higher number of optimization variables and active constraints increases the calculation speed of an active set solver significantly. Nevertheless, maximum computation times for the \gls{mpSQP} of around \SI{0.5}{\second} on the performance line exclude the \gls{QP} solver \gls{qpOASES} for this type of application.
\begin{table}[!tb]
	\renewcommand{\arraystretch}{1.1}
	\caption{Solver mean runtimes.}
	\label{tab:runtimes}
	\centering
	\begin{tabular}{|l c c c c|}
		\hline
		CPU & \multicolumn{3}{c}{Intel i7-7820HQ} & A57 ARM\\
		Prob. formulation & mpSQP & NLP & mpSQP & mpSQP\\
		Solver & OSQP & IPOPT & qpOASES & OSQP\\
		Performance in \SI{}{\milli\second} & 6.20 & 40.4 & 164 & 32.4\\
		Emergency in \SI{}{\milli\second} & 6.95 & 29.8 & 26.1 & 34.2\\
		\hline
	\end{tabular}
\end{table}
\section{Conclusion}
In this paper we presented a tailored \gls{mpSQP} algorithm capable of adaptive velocity planning in real time for race cars operating at the limits of handling, and velocities above \SI{200}{\kilo\meter\per\hour}. The planner can deal with performance and emergency velocity profiles. Furthermore, the optimization handles multi-parametric input, e.g., from the friction estimation module or the race \gls{ES}. We also specified the boundaries of maximum variation within these parameters to keep the problem feasible. Additionally, we compared different solvers applied to our problem formulation to compare calculation times as well as the solution qualities. Here, our \gls{mpSQP} in combination with the \gls{ADMM} solver \gls{OSQP} outperformed the active set strategy \gls{qpOASES} and the general \gls{NLP} solver \gls{IPOPT} in terms of calculation time, but reached nearly the same solution quality as \gls{IPOPT}. This indicates that the first order \gls{ADMM} in \gls{OSQP} shows its strength for the
minimum-time optimization problem as it handles the large set of active constraints well.

In future work we will apply the presented algorithm in autonomous races and implement a tailored trajectory optimization module based on the presented results and techniques for comparison. 


%



\section*{Contributions \& Acknowledgments}
T. H. initiated the idea of the paper and contributed
significantly to the concept, modeling, implementation and results. A. W. contributed to the design and feasibility analysis of the optimization problem. L. H. contributed essentially to the integration of variable acceleration limits. J. B. contributed to the whole concept of the paper.
M. L. provided a significant contribution to the concept
of the research project. He revised the paper critically for
important intellectual content. M. L. gave final approval for
the publication of this version and is in agreement with all
aspects of the work. As a guarantor, he accepts responsibility
for the overall integrity of this paper.

We would like to thank the Roborace team for giving us the opportunity to work with them and for the use of their vehicles for our research project. We would also like to thank the Bavarian Research Foundation (Bayerische Forschungsstiftung) for funding us in connection with the ``rAIcing'' research project. This work was also conducted with basic research fund of the Institute of Automotive Technology from the Technical University of Munich.

\ifCLASSOPTIONcaptionsoff
  \newpage
\fi



\bibliographystyle{IEEEtran}
\bibliography{IEEEabrv,referencesBIBTEX}
%
%
%

%

\begin{IEEEbiography}[{\includegraphics[width=1in,height=1.25in,clip,keepaspectratio]{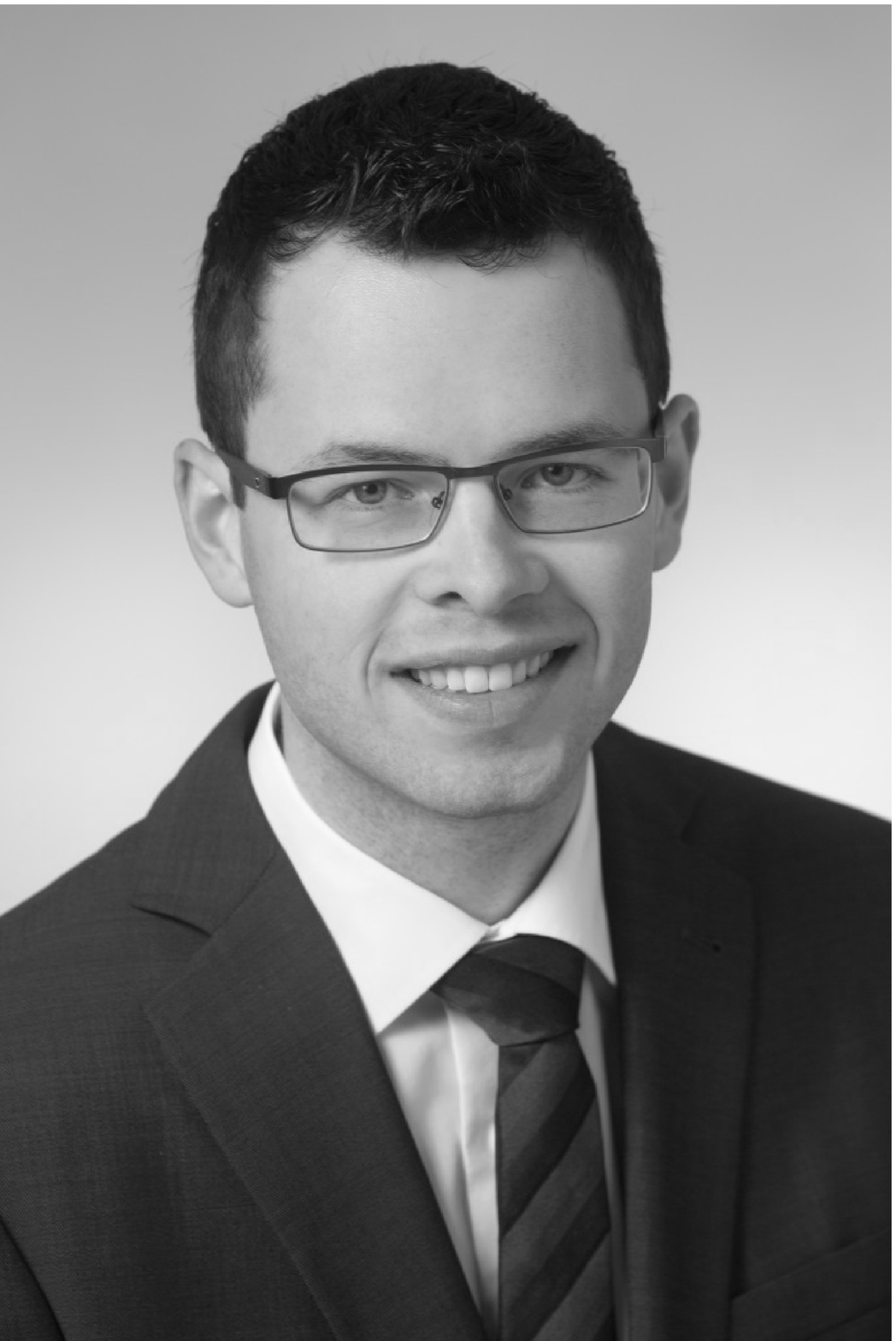}}]{Thomas Herrmann}
	was awarded a B.Sc. and an M.Sc. in Mechanical Engineering by the Technical University of Munich (TUM), Germany, in 2016 and 2018, respectively. He is currently pursuing his doctoral studies at the Institute of Automotive Technology at TUM where he is working as a Research Associate. His current research interests include optimal control in the field of trajectory planning for autonomous vehicles and the efficient incorporation of the electric powertrain behavior within these optimization problems.
\end{IEEEbiography}

\begin{IEEEbiography}[{\includegraphics[width=1in,height=1.25in,clip,keepaspectratio]{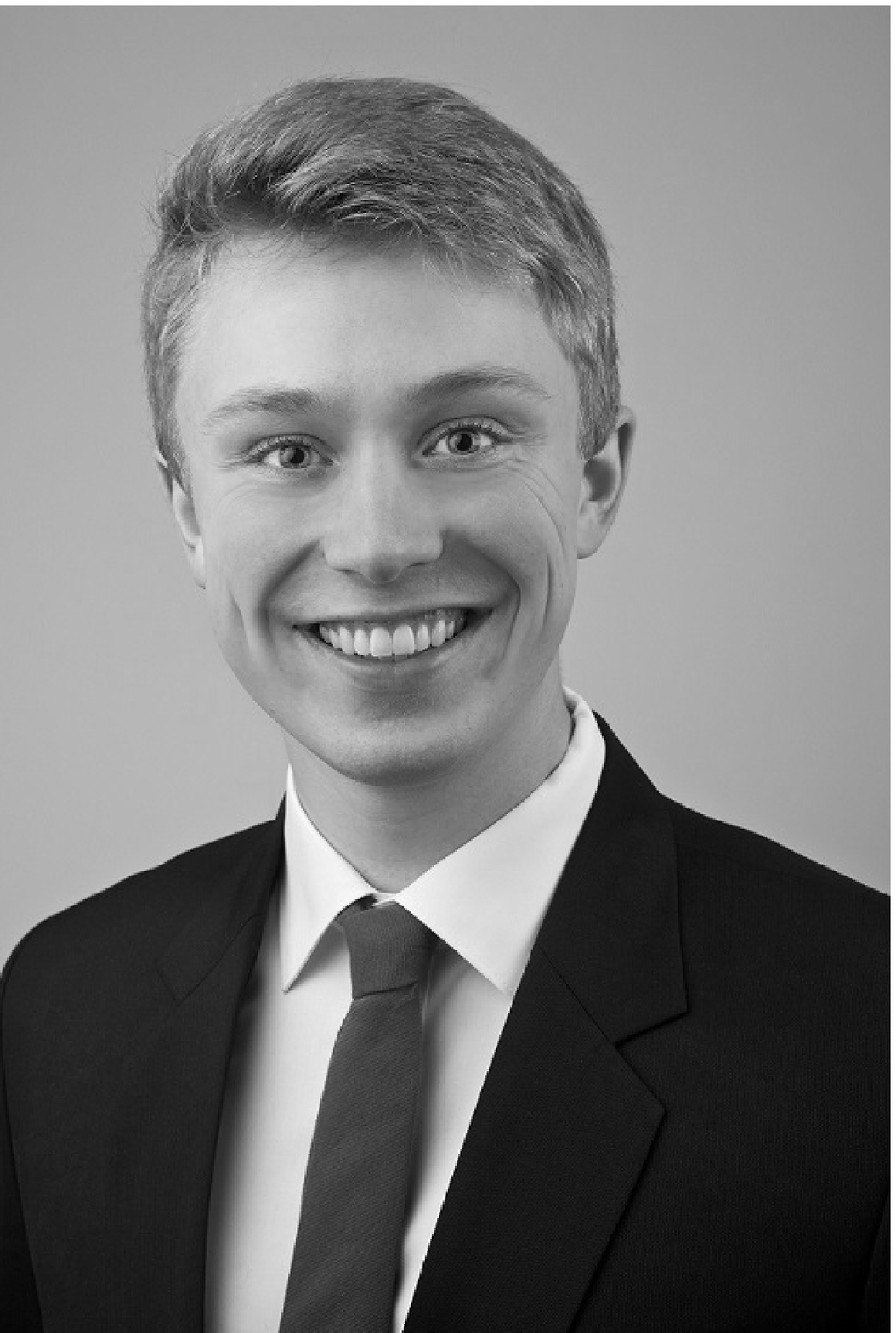}}]{Alexander Wischnewski}
	was awarded a B.Eng. in Mechatronics by the DHBW Stuttgart in 2015, and an M.Sc. in Electrical Engineering and Information Technology by the University of
	Duisburg-Essen, Germany, in 2017. He is currently pursuing his doctoral studies at the Chair of Automatic Control at the Department of Mechanical Engineering at Technical University of Munich (TUM), Germany. His research interests lie at the intersection of control engineering and machine learning, with a strong focus on autonomous driving. 
\end{IEEEbiography}

\begin{IEEEbiography}[{\includegraphics[width=1in,height=1.25in,clip,keepaspectratio]{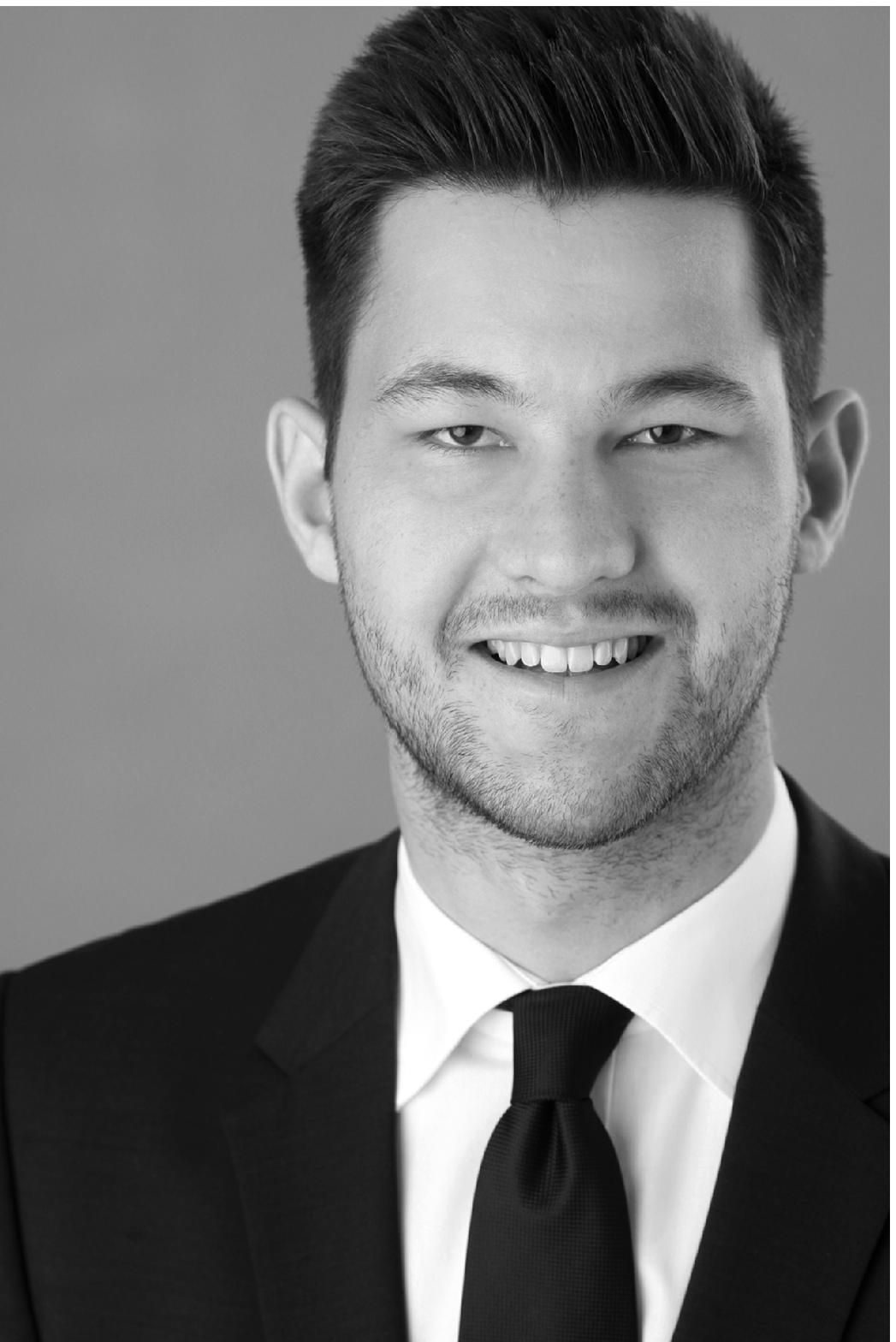}}]{Leonhard Hermansdorfer}
	was awarded a B.Sc. and an M.Sc. in Mechanical Engineering by 
	the Technical University of Munich (TUM), Germany, in 2015 and 2018, respectively. He is currently pursuing his doctoral studies at the Institute of Automotive Technology at TUM where he is working as a Research Associate. His current research interest is the identification of a vehicle's maximum transmittable tire forces, with a strong focus on model-less approaches based on machine learning.
\end{IEEEbiography}

\begin{IEEEbiography}[{\includegraphics[width=1in,height=1.25in,clip,keepaspectratio]{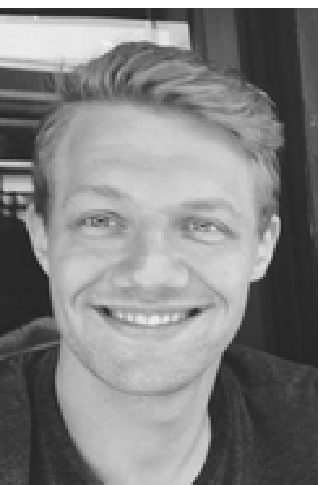}}]{Johannes Betz}
	is currently a postdoctoral researcher at the Technical University of Munich (TUM). He graduated in 2013 with an M.Sc. in Automotive Technology from the University of Bayreuth, Germany. He graduated with a doctoral degree from TUM about the topic of fleet disposition for electric vehicles. Since then, he continues his research in the field of autonomous driving. His research topics include the dynamic trajectory and behavioral planning for autonomous vehicles at the handling limits, as well as prediction of traffic participant behavior in non-deterministic environments.
\end{IEEEbiography}

\begin{IEEEbiography}[{\includegraphics[width=1in,height=1.25in,clip,keepaspectratio]{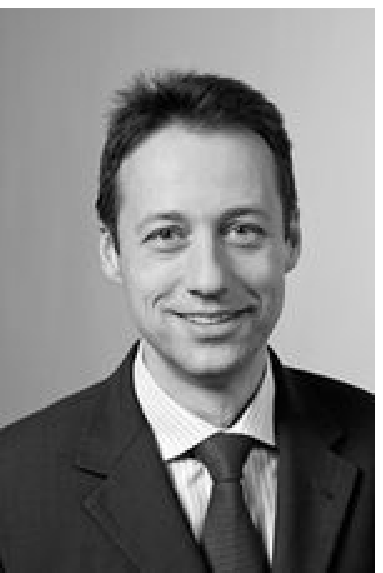}}]{Markus Lienkamp}
	studied Mechanical Engineering at TU Darmstadt, Germany, and Cornell University, USA, and received a Ph.D. degree from TU Darmstadt in 1995. He was with Volkswagen as part of an International Trainee Program and took part in a joint venture between Ford and Volkswagen in Portugal. Returning to Germany, he led the Brake Testing Department, VW Commercial Vehicle Development Section, Wolfsburg. He later became the Head of the ``Electronics and Vehicle'' Research Department, Volkswagen AG’s Group Research Division. His main priorities were advanced driver assistance systems and vehicle concepts for electromobility. He has been Head and Professor of the Institute of Automotive Technology, Technical University of Munich (TUM), Germany, since November 2009. He is conducting research in the area of electromobility, autonomous driving and mobility.
\end{IEEEbiography}






\end{document}